\documentclass[10pt,journal,compsoc]{IEEEtran}
\IEEEoverridecommandlockouts

\usepackage[acronym]{glossaries}
\usepackage{xcolor}
\usepackage{graphicx}
\usepackage{amsmath}
\usepackage{mathtools}
\usepackage{amssymb}
\usepackage{threeparttable}
\usepackage{enumitem}
\usepackage{multirow}
\usepackage{siunitx}
\usepackage{tabulary}
\usepackage{booktabs}
\usepackage[frozencache=true,cachedir=MintedCache]{minted}
\usepackage{comment}
\usepackage[nocompress]{cite}
\usepackage{listings}
\usepackage{afterpage}
\usepackage{caption}
\usepackage{lettrine}
\usepackage{tikz}
\usepackage{pifont}
\usepackage{tablefootnote}
\usepackage{hyperref}

\definecolor{darkspringgreen}{rgb}{0.09, 0.45, 0.27}
\definecolor{checkmarkboxgreen}{rgb}{0.20, 0.69, 0.2}
\definecolor{crossmarkboxred}{rgb}{0.80, 0.2, 0.2}
\definecolor{springgreen}{rgb}{0.26, 0.70, 0.30}

\newcommand*\circled[1]{\tikz[baseline=(char.base)]{
            \node[shape=circle,fill,inner sep=1pt] (char) {\textcolor{white}{#1}};}}

\DeclareRobustCommand*\circledGreen[1]{\tikz[baseline=(char.base)]
            \node[shape=circle,fill=springgreen,inner sep=1pt] (char) {\textcolor{white}{#1}};}

\newcommand{\circledcheckmark}{
    \raisebox{-5pt}{
    \begin{tikzpicture}
        \node[fill = checkmarkboxgreen, circle, inner sep=0.3pt, text=white, outer sep=0pt] (number) at (0, 0, 0) {\footnotesize{{\fontsize{9pt}{18pt}\selectfont \ding{51}}}};
    \end{tikzpicture}
    }
}

\newcommand{\circledcrossmark}{
    \raisebox{-5pt}{
    \begin{tikzpicture}
        \node[fill = crossmarkboxred, circle, inner sep=1.0pt, text=white, outer sep=0pt] (number) at (0, 0, 0) {\footnotesize{{\fontsize{9pt}{18pt}\selectfont \ding{55}}}};
    \end{tikzpicture}
    }
}

\newcommand{\Rmnum}[1]{\expandafter\@slowromancap\romannumeral #1@}

\newlist{shortletterenum}{enumerate}{10}
\setlist[shortletterenum]{label*=\alph*.,nosep,leftmargin=*}
\newlist{shortenum}{enumerate}{10}
\setlist[shortenum]{label*=\arabic*.,nosep,leftmargin=*}
\newlist{shortitem}{itemize}{10}
\setlist[shortitem]{label*=-,nosep,leftmargin=*}

\newacronym{AI}{AI}{Artificial Intelligence}
\newacronym{DL}{DL}{Deep Learning}
\newacronym{DNN}{DNN}{Deep Neural Network}
\newacronym[longplural={Application-Specific Integrated Circuits}]{ASIC}{ASIC}{Application-Specific Integrated Circuit}
\newacronym[longplural={Graphics Processing Units}]{GPU}{GPU}{Graphics Processing Unit}
\newacronym{TPU}{TPU}{Tensor Product Unit}
\newacronym{CPU}{CPU}{Central Processing Unit}
\newacronym[longplural={Instruction Set Architectures}]{ISA}{ISA}{Instruction Set Architecture}
\newacronym[plural=TCNs, firstplural={Temporal Convolutional Neural Networks (TCNs)}]{TCN}{TCN}{Temporal Convolutional Neural Network}
\newacronym[plural=CNNs, firstplural={Convolutional Neural Networks (CNNs)}]{CNN}{CNN}{Convolutional Neural Network}
\newacronym{MHSA}{MHSA}{Multi-Head Self-Attention}
\newacronym{MHA}{MHA}{Multi-Head Attention}
\newacronym[longplural={Reduced Instruction Set Computers}]{RISC}{RISC}{Reduced Instruction Set Computer}
\newacronym{ARM}{ARM}{Advanced RISC Machine}
\newacronym{SSR}{SSR}{Stream Semantic Register}
\newacronym{HW}{HW}{Hardware}
\newacronym{WL}{WL}{Workload}
\newacronym{ONNX}{ONNX}{Open Neural Network Exchange}
\newacronym{NAS}{NAS}{Neural Architecture Search}
\newacronym{PULP}{PULP}{Parallel Ultra Low Power}
\newacronym{AXI}{AXI}{Advanced eXtensible Interface}
\newacronym{TCDM}{TCMD}{Tightly Coupled Data Memory}
\newacronym[longplural={Micro-Controller Units}]{MCU}{MCU}{Micro-Controller Unit}
\newacronym{ODL}{ODL}{On-Device Learning}
\newacronym{SGD}{SGD}{Stochastic Gradient Descent}
\newacronym{FLOPS}{FLOPS}{Floating Point Operations Per Second}
\newacronym{SNN}{SNN}{Spiking Neural Network}
\newacronym{KWS}{KWS}{Keyword Spotting}
\newacronym{IIS}{IIS}{Integrated Systems Laboratory}
\newacronym[longplural={Large Language Models}]{LLM}{LLM}{Large Language Model}
\newacronym[longplural={Systems-on-Chip}]{SoC}{SoC}{System-on-Chip}
\newacronym{NLP}{NLP}{Natural Language Processing}
\newacronym{SotA}{SotA}{State of the Art}
\newacronym{CV}{CV}{Computer Vision}
\newacronym{MAC}{MAC}{Multiply-Accumulate}
\newacronym{IoT}{IoT}{Internet of Things}
\newacronym{SIMD}{SIMD}{Single Instruction Multiple Data}
\newacronym{PTQ}{PTQ}{Post-Training Quantization}
\newacronym{QAT}{QAT}{Quantization Aware Training}
\newacronym{EEG}{EEG}{Electroencephalogram}
\newacronym{RAW}{RAW}{Read-After-Write}
\newacronym{WRL}{WRL}{Weight-Reuse Linear}
\newacronym{IRL}{IRL}{Input-Reuse Linear}
\newacronym{LWT}{LWT}{Layer-Wise Tiling}
\newacronym{DFT}{DFT}{Depth-First Tiling}
\newacronym{MQA}{MQA}{Multi-Query Attention}
\newacronym{GQA}{GQA}{Grouped-Query Attention}
\newacronym[longplural={Vision Transformers}]{ViT}{ViT}{Vision Transformer}
\newacronym{GELU}{GELU}{Gaussian Error Linear Unit}
\newacronym{GEMM}{GEMM}{GEneral Matrix Multiplication}
\newacronym{HPC}{HPC}{High-Performance Computing}
\newacronym[longplural={Direct Memory Accesses}]{DMA}{DMA}{Direct Memory Access}
\newacronym{PE}{PE}{Processing Element}
\newacronym{FPU}{FPU}{Floating-Point Unit}
\newacronym{RAM}{RAM}{Random-Access Memory}
\newacronym{SRAM}{SRAM}{Static Random-Access Memory}
\newacronym{NE16}{NE16}{Neural Engine 16-channels}
\newacronym{FWSA}{FWSA}{Fused-Weight Self-Attention}
\newacronym{TQT}{TQT}{Trained Quantization Threshold}
\newacronym{ReLU}{ReLU}{Rectified Linear Unit}
\newacronym{ML}{ML}{Machine Learning}
\begin{document}
\title{Optimizing the Deployment of Tiny Transformers on Low-Power MCUs}
\vspace{-0.6cm}

\author{Victor~J.B.~Jung,
        Alessio~Burrello,
        Moritz Scherer,
        Francesco~Conti, 
        Luca~Benini
\IEEEcompsocitemizethanks{
\IEEEcompsocthanksitem Victor~J.B.~Jung, Moritz Scherer, L. Benini are with the Integrated Systems Laboratory (IIS) of ETH Z\"urich, ETZ, Gloriastrasse 35, 8092 Z\"urich, Switzerland (e-mail: name.surname@iis.ee.ethz.ch).
\IEEEcompsocthanksitem A. Burrello, F. Conti, L. Benini are with the Department of Electrical, Electronic and Information Engineering, University of Bologna, 40136 Bologna, Italy.
E-mail: firstname.firstsurname@unibo.it
\IEEEcompsocthanksitem A. Burrello is also with the Interuniversity Department
of Regional and Urban Studies and Planning, Politecnico di Torino, 10129, Turin, Italy. E-mail: name.surname@polito.it
\IEEEcompsocthanksitem This work has received funding from the Chips Joint Undertaking (Chips-JU) TRISTAN project under grant agreement No 101095947. The JU receives support from the European Union’s Horizon Europe research and innovation program. Pre-print manuscript submitted for review to the IEEE Transactions on Computers.
}
\thanks{Manuscript received January XX, XXXX; revised January XX, XXXX.}}


\IEEEtitleabstractindextext{
\begin{abstract}
Transformer networks are rapidly becoming \gls{SotA} in many fields, such as \gls{NLP} and \gls{CV}.
Similarly to \glspl{CNN}, there is a strong push for deploying Transformer models at the extreme edge, ultimately fitting the tiny power budget and memory footprint of \glspl{MCU}.
However, the early approaches in this direction are mostly ad-hoc, platform, and model-specific.
This work aims to enable and optimize the flexible, multi-platform deployment of encoder Tiny Transformers on commercial \glspl{MCU}.
We propose a complete framework to perform end-to-end deployment of Transformer models onto single and multi-core \glspl{MCU}. 
Our framework provides an optimized library of kernels to maximize data reuse and avoid unnecessary data marshaling operations into the crucial attention block. A novel \gls{MHSA} inference schedule, named \gls{FWSA}, is introduced, fusing the linear projection weights offline to further reduce the number of operations and parameters. Furthermore, to mitigate the memory peak reached by the computation of the attention map, we present a \gls{DFT} scheme for \gls{MHSA} tailored for cache-less \gls{MCU} devices that allows splitting the computation of the attention map into successive steps, never materializing the whole matrix in memory. 
We evaluate our framework on three different \gls{MCU} classes exploiting ARM and RISC-V \gls{ISA}, namely the STM32H7 (ARM Cortex M7), the STM32L4 (ARM Cortex M4), and GAP9 (RV32IMC-XpulpV2).
We reach an average of 4.79\,$\times$ and 2.0\,$\times$ lower latency compared to \gls{SotA} libraries CMSIS-NN (ARM) and PULP-NN (RISC-V), respectively. Moreover, we show that our \gls{MHSA} depth-first tiling scheme reduces the memory peak by up to 6.19\,$\times$, while the fused-weight attention can reduce the runtime by 1.53\,$\times$, and number of parameters by 25\,\%. 
Leveraging the optimizations proposed in this work, we run end-to-end inference of three \gls{SotA} Tiny Transformers for three applications characterized by different input dimensions and network hyperparameters. 
We report significant improvements across the networks: for instance, when executing a transformer block for the task of radar-based hand-gesture recognition on GAP9, we achieve a latency of \SI{0.14}{ms} and energy consumption of \SI{4.92}{\micro\joule}, 2.32\,$\times$ lower than the \gls{SotA} PULP-NN library on the same platform. 
\end{abstract}

\glsresetall{}

}
\maketitle
\IEEEdisplaynontitleabstractindextext
\vspace{-0.6cm}

\IEEEraisesectionheading{\section{Introduction}}
\label{sec:intro}
\noindent
\lettrine[]{I}{n} the last few years, there has been a significant trend towards moving computing workload from centralized facilities towards the extreme edge of the \gls{IoT}, improving privacy, efficiency, and ensuring a predictable and constant latency, independent from the network coverage and congestion~\cite{banbury2021tinyMLBenchmarking}. As a consequence, modern \gls{IoT} endpoints have been changing from simple \glspl{MCU} equipped with the sensor infrastructure for data collection and transmission towards more complex heterogeneous \glspl{SoC} characterized by dedicated accelerators and enhanced \glspl{ISA} to optimize efficient on-board computations -- while still meeting to tight power, performance, and cost constraints. 
For this reason, extensive research has been conducted on how to squeeze complex \gls{ML} models into such constrained devices with a strong emphasis on \glspl{DNN} due to their success in solving many challenging tasks. 
Standard hardware-agnostic optimization techniques include topological changes~\cite{lin2020mcunet}, data quantization~\cite{jain2020tqt, gong2022VecQ}, and pruning~\cite{han2016deepC}. These efforts are often collectively labeled as \textit{TinyML} and are applied to \gls{SotA} \glspl{CNN}, such as ResNets~\cite{he2015resnet} or MobileNets~\cite{howard2017mobilenets}, that dominates most of the tasks of the \gls{IoT} domain, including image recognition, object detection, or time-series analysis.

At the same time, increasing efforts have been dedicated to finding a successor architecture to \glspl{CNN} for \gls{IoT} processing to increase tasks' accuracy further.
A strong candidate is the Transformer architecture~\cite{vaswani2017attention}, which first emerged as \gls{SotA} model in \gls{NLP} and is now also \gls{SotA} in other domains such as \gls{CV} and Audio processing. 
The Transformers architecture relies on the attention mechanism that, thanks to its generality and flexibility, has enabled significant advancements in language understanding, text generation, and translation tasks~\cite{lin2021surveyTr}. 
Transformer models have been initially developed targeting "at scale" deployment in the cloud. Industry and academia are now increasing efforts toward Transformer models tuned for edge deployment.
Recent advancements have focused on the minimization of Transformer model size~\cite{burrello2021mcuIsAllYouNeed} while still maintaining leading-edge accuracy when compared with \gls{CNN} models of similar size, thus introducing a novel class of Tiny Transformers.
The success of these early reduced-scale Transformer models confirms that Tiny Transformers are relevant and applicable in Edge computing scenarios.  

However, the introduction of Transformers comes with novel computational primitives not commonly found in \glspl{CNN} and has not been optimized so far. Precisely, the \gls{MHSA} operation comes with many challenges, such as the high memory footprint of intermediate results and frequent data marshaling.
Despite our early work demonstrating encoder-only Tiny Transformer deployment~\cite{burrello2021mcuIsAllYouNeed}, the approach was highly model and platform-specific and not easily generalizable to multiple models and platforms. Moreover, early work did not tackle the critical problem of optimized data tiling and computation scheduling, which is essential for most practical Tiny Transformers. 

To facilitate the deployment and the optimization of a wide range of encoder-only Transformer models for TinyML platforms on multiple \gls{MCU} platforms, we significantly extend an existing deployment framework (DORY~\cite{burrello2021dory}) and quantization library (Quantlib).
Thus, we present a comprehensive end-to-end and open-source\,\footnote{\url{https://github.com/pulp-platform/pulp-transformer}} encoder-only Transformer deployment flow to \gls{SotA} \glspl{MCU}. 
Further, we explore optimizations at various levels of the stack, from pure \textit{algorithmic techniques}, to \textit{low-level code optimization}, \textit{novel scheduling} and \textit{tiling schemes}. Together, these optimizations significantly enhance the performance of Transformers deployment on low-power \glspl{MCU}. In detail, we present three main contributions:

\begin{enumerate}

\item We designed a library of highly efficient kernels for both \gls{MHSA} and \gls{FWSA}. The library is tailored to minimize the overhead associated with data marshaling operations and maximize data reuse through carefully optimized data layout and loop reordering for each Attention layer. By reducing unnecessary data movements and optimizing memory access patterns, the proposed library significantly boosts the overall performance of Attention in an \gls{ISA}-agnostic fashion, with comparable improvements for RISC-V and ARM with respect to \gls{SotA} \gls{DNN} libraries PULP-NN and CMSIS-NN.

\item We introduce two novel optimizations: a fusion-based tiling scheme for \gls{MHSA} operations and an offline weight fusion schedule for the \gls{MHSA} operation. By combining fusion-based tiling and pipelining-related computations, the memory footprint and frequency of memory transfers during \gls{MHSA} computation are substantially diminished, reducing the memory peak to 6.19\,$\times$. Furthermore, the weight fusion schedule reduces the latency by a factor of 1.53$\times$ while reducing the number of parameters by 25\,\%.  

\item We extensively benchmark our library and optimizations using published transformers targeting three tasks, i.e., seizure detection from EEG signals, arrhythmia classification from ECG signals, and hand-gesture recognition from high-frequency short-range pulsed RADAR. 
\end{enumerate}

\noindent
Experimental evaluations on the above-mentioned \glspl{MCU} demonstrate an energy consumption reduction compared to SoA libraries of up to 5.1\,$\times$ on GAP9 and 2.9\,$\times$ on ARM-based platforms when executing the attention-building block of the three transformers. 
When considering end-to-end execution of the networks on GAP9, we obtain the best latency of \SI{9.42}{\ms}, \SI{2.85}{\ms}, and  \SI{5.49}{\ms} for the three different tasks at the maximum frequency of 370 MHz. In the most energy-efficient configuration, we achieved \SI{310}{\micro\joule}, \SI{90}{\micro\joule}, \SI{207}{\micro\joule} energy consumption, respectively, still respecting the real-time constraint of the applications.

The rest of the paper is structured as follows. Sec.~\ref{sec:background} provides the necessary background while Sec.~\ref{sec:related} summarizes related works on \gls{MHSA} optimization and TinyML platforms. In Sec.~\ref{sec:method}, we provide implementation details of the library, tiling scheme, and Weight-Fusion schedule. Finally, Sec.~\ref{sec:results} presents experimental results, and Sec.~\ref{sec:conclusion} concludes the paper.

\section{Background}
\label{sec:background}
\subsection{Transformers and Multi-Head Self-Attention}

\begin{figure}[t]
    \centering
    \includegraphics[width=0.8\linewidth]{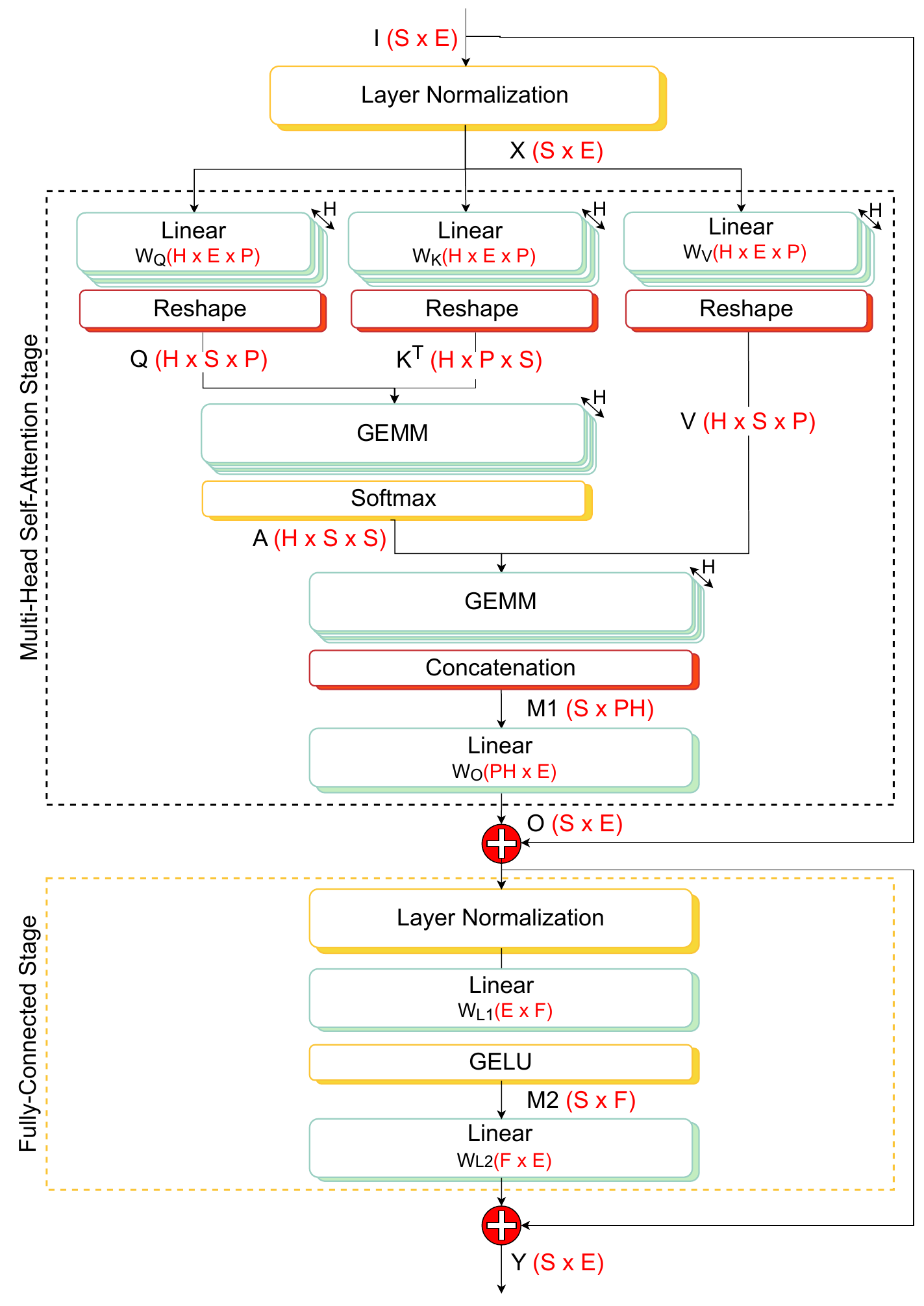}
    \caption{Topology of the Transformer block, composed of a \gls{MHSA} stage and a Fully-Connected stage. The dimensions of the tensors are indicated in red.}
    \label{fig:attention}
\end{figure}

Recently, transformer-based architectures have had increasing success in various domains thanks to their flexibility, high accuracy, and generalization capabilities. More precisely, encoder-decoder~\cite{raffel2023t5} and decoder-only~\cite{touvron2023llama} models are now dominating the field of \gls{NLP} while encoder-only~\cite{devlin2018bert} is widely used in \gls{CV}.
In particular, models from tens of millions to tens of billions of parameters have been designed to solve complex problems such as question-answering and image generation~\cite{devlin2018bert, ramesh2022dalle2}.
On the other side, in the tinyML field, new encoder-only transformer networks are emerging, demonstrating excellent capability, especially in time-series analysis~\cite{busia2022eegformer}. Transformers are built around repeating an identical Transformer block. 
As shown in Figure \ref{fig:attention}, the Transformer block comprises two stages, the \gls{MHSA} and the Fully-Connected stage. To deploy transformers on \glspl{MCU}, the \gls{MHSA} stage is the most challenging step, given that the Fully-Connected stage is common in \glspl{CNN} and its optimization has already been tacked in previous works.

The dimensions of each operation in the \gls{MHSA} are determined by four parameters: the \textit{sequence length} $S$, the \textit{embedding dimension} $E$, the \textit{projection dimension} $P$ and the \textit{head dimension} $H$.
The first step in the \gls{MHSA} stage is the projection of the input sequence $X \in \mathbb{R}^{S\times E}$ into the query, key, and values, $Q, K, V \in \mathbb{R}^{S\times P}$, by means of three weights matrices, $W_\text{query}, W_\text{key}, W_\text{value} \in \mathbb{R}^{H\times E\times P}$.
\begin{equation}
    \mathbf{Q} = \mathbf{X}\mathbf{W}_\text{query}, \qquad
    \mathbf{K} = \mathbf{X}\mathbf{W}_\text{key},   \qquad
    \mathbf{V} = \mathbf{X}\mathbf{W}_\text{value}.
\label{eq:linear}
\end{equation}
In the second step, $Q$, $K$, and $V$ are combined in the \textit{Attention} step, which is the core of the Transformer block. It is defined as
\begin{equation}
    \text{Attention}(\mathbf{Q}, \mathbf{K}, \mathbf{V})
    \coloneqq
    \text{Softmax} \left(\frac{\mathbf{Q}\mathbf{K}^\text{T}}{\sqrt{d}} \right)\mathbf{V},
\label{eq:attention}
\end{equation}
with $d$ being the dimensionality of $\mathbf{K}$ used as a scaling factor and where the softmax function is applied to each row. The softmax of the $i$-$th$ element of a row of size $n$ is defined as follows:
\begin{equation}
    \text{Softmax}(\boldsymbol{x})_i = \frac{e^{x_i-\max(\boldsymbol{x})}}{\sum_{j=1}^{n} e^{x_j-\max(\boldsymbol{x})}}
\end{equation}
The output of this second step is the matrix $M1$.
These first two steps are repeated for every head of the \gls{MHSA} layer, leading to $H$ parallel executions of the three Linear projection layers and the two \glspl{GEMM} of Eq.~\ref{eq:linear} and Eq.~\ref{eq:attention}.
The final step of the \textit{\gls{MHSA}} stage involves a Linear layer to project the matrix $M1$ back into the input embedding space $E$, fusing the computation from the different heads. 

The second stage of the Transformer block, shown in Figure~\ref{fig:attention}, is the \textit{Fully-Connected} stage. It is an entirely sequential stage comprising a row-wise layer normalization, an element-wise \gls{GELU}~\cite{hendrycks2023gelu}, and two Linear layers.

In Table~\ref{tab:attention_workload}, we report the number of parameters and the computational workload for the two stages in Figure~\ref{fig:attention}. To measure the workload's complexity, we account for the number of \gls{MAC} operations in \glspl{GEMM} and Linear layers. However, it is essential to note that other operators, such as softmax, transposition, concatenation, and \gls{GELU}, can also strongly impact overall latency.
\begin{table}[]
\caption{Number of parameters and operations for the \textit{\gls{MHSA}} and \textit{Fully-Connected} stage described in Fig \ref{fig:attention}.}
\centering
\renewcommand{\arraystretch}{1.35}
\label{tab:attention_workload}
\begin{tabular}{@{}l|ll@{}}
\hline
                & Parameters         & MAC            \\ \hline
\gls{MHSA}      & $4SEPH$            & $2SPH(2E+S)$  \\
Fully-Connected & $2FE$             & $2S(FE+2F+E)$          \\ \hline
\end{tabular}
\end{table}

\subsection{Hardware Platforms: RISC-V and ARM MCUs}
\label{subsec:hardware}

We consider platforms relevant to the TinyML context, typically featuring $\sim$\,\SI{1}{\mega\byte} of on-chip memory~\cite{banbury2021tinyMLBenchmarking} and based on the two leading \SI{32}{\bit} \glspl{ISA} for \glspl{MCU}, namely the ARMv7 and RISC-V \SI{32}{\bit} (RV32).
We open-source our framework and target commercially available \glspl{MCU} to facilitate benchmarking and extensions. Specifically, we report results on two single-core ARM \glspl{MCU}, based on low-power Cortex-M4 and high-performance Cortex-M7 cores, as well as one multi-core RISC-V \gls{MCU}, to span a broad spectrum of performance in the \gls{MCU} market.

\subsubsection{ARM Cortex platforms}
\textbf{STM32L4 - Cortex-M4\,\footnote{\url{https://developer.arm.com/documentation/dui0553}}}: The STM32L4 has a Cortex-M4 core coupled with \SI{640}{\kilo\byte} of \gls{RAM} and \SI{2}{\mega\byte} of Flash. 
Among the platforms chosen, the M4 shows the lowest average power consumption of \SI{13.63}{\mW} at a clock frequency of \SI{80}{\mega\hertz}. 
Additionally, the core supports \SI{16}{\bit} \gls{SIMD} operations. In detail, the board used is the STM32L476\,\footnote{\url{https://www.st.com/resource/en/datasheet/stm32l476je.pdf}}.

\textbf{STM32H7 - Cortex-M7\,\footnote{\url{https://developer.arm.com/documentation/ddi0489}}}: This architecture is representative of the high-performance end of the spectrum for ARM \glspl{MCU}. At a frequency of \SI{480}{\mega\hertz}, it consumes \SI{234}{\mW} on average. Among the additional features, we highlight a more complex memory hierarchy involving both data and instruction caches as well as a \gls{DMA} for software-managed memory movements between the main memory (\SI{1}{\mega\byte} of \gls{RAM}) and a smaller but faster data memory of \SI{64}{\kilo\byte}. Specifically, in this work, we use the NUCLEO-H743ZI2 board\,\footnote{\url{https://www.st.com/en/evaluation-tools/nucleo-h743zi.html}}.

We do not target hardware platforms using the ARM Helium vector extension\,\footnote{\url{https://developer.arm.com/documentation/102102}}, such as the Cortex-M55\,\footnote{\url{https://developer.arm.com/documentation/101051}} and Cortex-M85\,\footnote{\url{https://developer.arm.com/documentation/101924}}, given that they are not yet widely available on the market.

\subsubsection{RISC-V platform: GAP9}
GAP9\,\footnote{\url{https://greenwaves-technologies.com/gap9_processor/}} is a low-power RISC-V multi-core processor commercialized by GreenWaves Technologies targeting \gls{DNN} workloads for hearable and smart sensors. This platform features a RISC-V control core for I/O management and a compute cluster. The cluster comprises 9 RISC-V cores to parallelize computational intensive workload and one convolution accelerator, named \gls{NE16}. Additionally, it includes four \glspl{FPU} as well as a square root and division unit. In this work, we focus on the RISC-V cores and do not use the NE16 accelerator because we want to keep our approach general such that it can be used in any \gls{MCU}.
The RISC-V cores present a 4-stages in-order single-issue pipeline based on the RV32 XpulpV2 \gls{ISA} extension~\cite{garofalo2020pulp}, tailored for fast and efficient signal processing. 
In particular, XpulpV2 includes features such as hardware loops, post-increment loads/stores, and \gls{SIMD} 8-bit instructions for quantized data. These extensions reduce execution time by more than $5\times$ for different workloads with respect to RV32 baseline~\cite{garofalo2020pulp}.

Every core in the cluster is tightly connected to an L1 memory of \SI{128}{\kilo\byte} via a single cycle logarithmic interconnect. The on-chip L2 memory communicates with the L1 through the \gls{AXI} bus, and its size is \SI{1.5}{\mega\byte}. To overlap computation with data transfers, GAP9 relies on two \gls{DMA} units to explicitly move data between L1 and L2 or between L2 and external memories. The \gls{DMA} core responsible for L1-L2 transfers reaches a peak bandwidth of \SI{2}{\giga\byte/\second}.

We do not include single-core RISC-V \glspl{MCU} to the benchmark, such as the ESP32-C3, as they use the vanilla RV32 \gls{ISA} and are dominated by the RV32-XpulpV2 cores in terms of performance and energy efficiency.

\begin{table*}[t]
\centering
\begin{tabular}{@{}lcccccc@{}}
\toprule
Library           & \begin{tabular}[c]{@{}c@{}}Supported \\ Platform\end{tabular} & Precision & \begin{tabular}[c]{@{}c@{}}Multi-Core \\ Support\end{tabular} & \begin{tabular}[c]{@{}c@{}}SIMD \\ Support\end{tabular} & \begin{tabular}[c]{@{}c@{}}Loop \\ Unrolling\end{tabular} & \begin{tabular}[c]{@{}c@{}}Fused Data \\ Marshalling\end{tabular} \\ \midrule

XNNPACK\,\footnotemark         & ARM/x86/RISC-V   &  fp16/fp32        &   \circledcrossmark   &   \circledcheckmark   &  \circledcheckmark  &   \circledcrossmark     \\[7pt]
QNNPACK\,\footnotemark                & ARM/x86          &  fp16/fp32        &   \circledcrossmark   &   \circledcheckmark   &  \circledcheckmark  &   \circledcrossmark     \\[7pt]
CMSIS-NN~\cite{lai2018cmsis}     & ARM              &  int8/int16       &   \circledcrossmark   &   \circledcheckmark   &  \circledcheckmark  &   \circledcrossmark     \\[7pt]
PULP-NN~\cite{garofalo2020pulp}  & RISC-V           &  int8             &   \circledcheckmark   &   \circledcheckmark   &  \circledcheckmark  &   \circledcrossmark     \\[7pt]
TinyFormer (Ours)                & RISC-V/ARM        &  int8             &   \circledcheckmark   &   \circledcheckmark   &  \circledcheckmark  &   \circledcheckmark     \\[5pt] \bottomrule
\end{tabular}
\caption{Comparison of the SotA kernel libraries for TinyML platforms.}
\label{tab:libraryComparison}
\end{table*}
\vspace{-1em}

\section{Related Work}
\label{sec:related}
\subsection{Attention Mechanism Optimizations}

Since the release of the seminal attention paper in 2017~\cite{vaswani2017attention}, the research community put considerable effort into optimizing the basic \gls{MHSA}, the major building block in transformer-based architectures introduced in the previous section. These optimizations can be classified into topology optimizations, software optimizations, and hardware acceleration.

\subsubsection{Topology optimizations}
The first class of optimizations aims to modify the attention mechanism's topology to reduce deployment costs regarding memory footprint or number of operations.

These approaches, including linearized attention, usually seek to eliminate the quadratic scaling of memory and computational requirements relative to the sequence length. To linearize models, researchers use either the kernel trick~\cite{katharopoulos2020tranAreRNN, choromanski2022performers, bolya2022hydra, peng2023rwkv} or low-rank approximation~\cite{wang2020linformer}. However, linear transformers suffer from performance degradation on various tasks compared to traditional attention~\cite{qin2022devil}. Hence, linear attention methods are currently not widely adopted by the research community~\cite{qin2022devil}. Consequently, this work focuses on speeding up the traditional attention mechanisms. 

Another popular topological change to the Attention mechanism is the \gls{MQA}~\cite{shazeer2019mqa}. It reformulates the \gls{MHSA} to use only one $K$ and $V$ head instead of $H$ heads. This modification drastically reduces the required memory bandwidth and increases data reuse. However, the impact on the output quality is non-negligible~\cite{ainslie2023gqa}. \gls{GQA} is a variant of \gls{MQA} proposed to reduce this accuracy drop. Instead of reducing the number of $K$ and $V$ heads to one, it groups a certain number of $KV$ heads, allowing the programmer to mitigate the accuracy drop. \gls{GQA} has been successful used in popular \glspl{LLM} such as LLama-2 ~\cite{touvron2023llama}. However, to the best of the authors' knowledge, these methods have yet to be successfully applied in the TinyML domain. Furthermore, the implementation of \gls{GQA} kernels can be derived from the basic ones and does not require extra optimization steps.

\footnotetext[10]{\url{https://github.com/google/XNNPACK}}
\footnotetext[11]{\url{https://github.com/pytorch/QNNPACK}}

\subsubsection{Software optimizations}
The second optimizations category encompasses every software optimization of the traditional \gls{MHSA} operator. One can notice that the vast majority of these works focus on large-scale inference using server-class hardware~\cite{dao2023flashattention, kwon2023paged}. 
Due to the extreme memory constraints and the small number of cores present in TinyML platforms, many fundamental differences arise in the way one would optimize the \gls{MHSA} operator. However, despite not being directly applicable to edge devices, most of these techniques can be used as starting points.

We excluded batch size-based optimizations, given that, due to the memory constraints of \glspl{MCU} and the context of online inference, we always consider batch sizes of 1 in this work.
Further, we do not consider transformer model "partitioning"~\cite{pope2022transcale}: While this is crucial for large transformer models that are too large for a single compute unit (\textit{e.g.} a \gls{GPU}), tinyML systems are overwhelmingly based on a single MCU.

Some works present optimizations partially applicable in the context of TinyML platforms~\cite{dao2023flashattention, shazeer2019mqa, ainslie2023gqa}. FlashAttention and FlashAttention-2 use online softmax and normalization~\cite{milakov2018online} to break the row dependencies, which results in better tiling. Together with carefully tuned kernels, FlashAttention significantly improves the performances of transformer inference of large models on \glspl{GPU} and is now fully integrated into the PyTorch library. However, some of the optimizations FlashAttention uses are not beneficial for extreme edge platforms, given the smaller memory and fewer computational resources. For instance, in our work, we favor output stationary tiling instead of the block tiling proposed in~\cite{dao2023flashattention}, alleviating the need to use the online softmax and normalization.
Additionally, many transformer models exhibit a high amount of sparsity. Therefore, various approaches exploiting this sparsity have been studied to accelerate the inference~\cite{liu2022DynamicSparseAtt}. However, sparsity exploitation generally requires dedicated hardware support to be beneficial, which is not available in the current generation of \glspl{MCU}. 

\subsubsection{Hardware accelerators}
Hardware accelerators optimize efficiency by tailoring architecture and circuits to specific computational patterns at the price of flexibility. Many have been proposed to accelerate \gls{MHSA} in all kinds of contexts, from server-class to the edge~\cite{ham2021elsa}. 
However, hardware accelerators usually employ a fixed dataflow and attention flavor, reducing the adaptability to different attention mechanisms (\textit{i.e.} such as \gls{GQA} or \gls{MQA}) and network topologies.
Their extra cost and limited flexibility limit their adoption in the current generation of \glspl{MCU}.
%
Hence, we focus on the most general and accessible way to port transformers to the edge: computing \gls{MHSA} on the ARM or RISC-V cores. 

\begin{figure}[t]
  \centering
    \includegraphics[trim={0cm 0cm 0cm 0cm},width=\columnwidth]{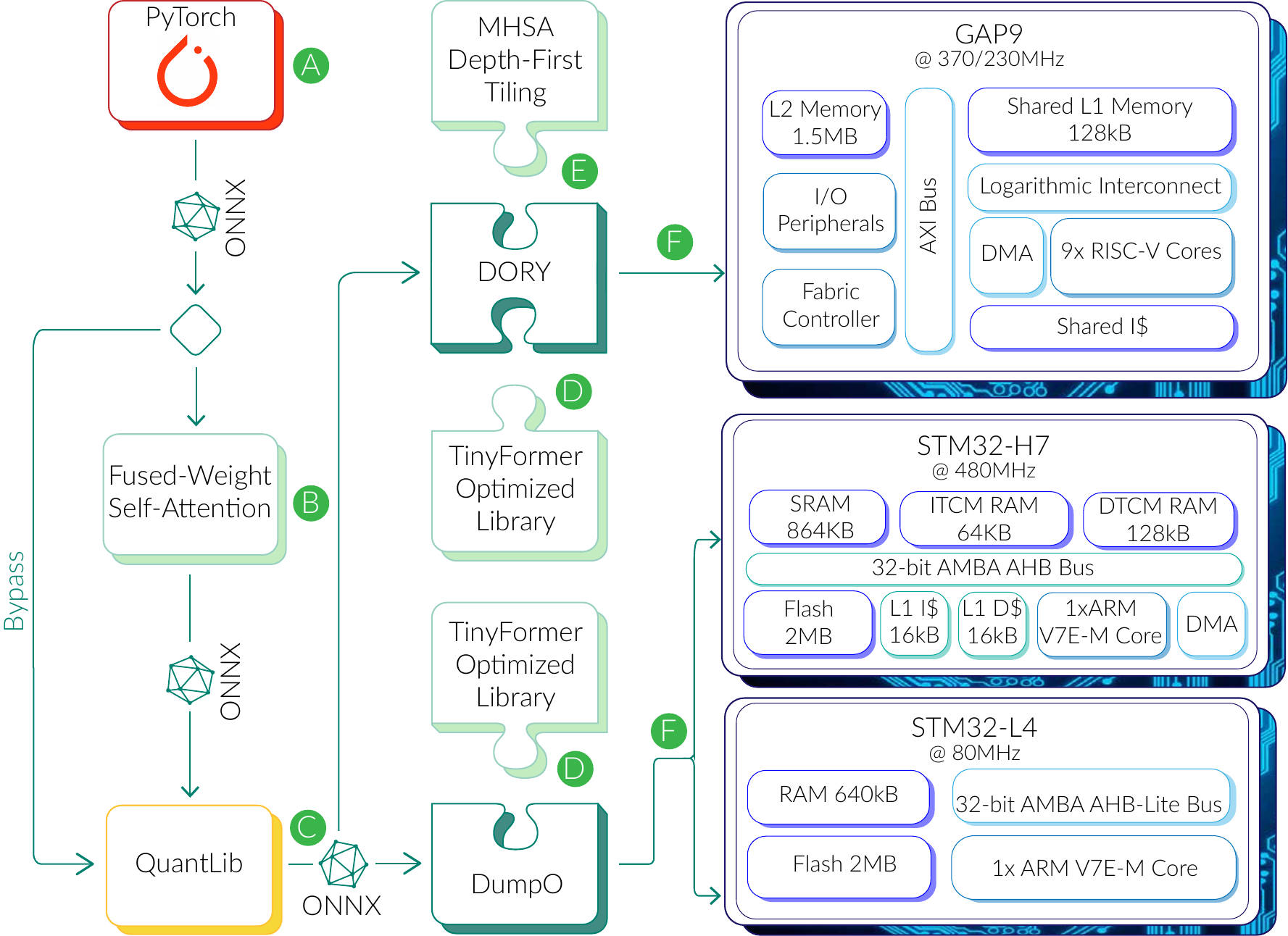}
  \caption{Overview of our Tiny Transformers deployment flow. The floating point Pytorch model in \protect\circledGreen{A} can be transformed by the \gls{FWSA} node in \protect\circledGreen{B} and is then fed to QuantLib \protect\circledGreen{C}. Afterward, the quantized \gls{ONNX} graph is ingested by the deployment frameworks enhanced with our library \protect\circledGreen{D} and \gls{DFT} optimization \protect\circledGreen{E}. Finally, in \protect\circledGreen{F}, the generated C code is deployed on the desired platforms.}
  \label{fig:flow}
\end{figure}

\vspace{-1em}
\subsection{DNN Kernel Libraries for TinyML Platforms}

Optimized kernel libraries are at the core of every \gls{DNN} deployment stack. These libraries contain manually optimized kernels to efficiently run operators commonly found in \glspl{DNN} such as convolutions or matrix multiplications.
Efficient kernel design relies heavily on hardware-specific expert knowledge and is very time-consuming. Despite some attempts of automatizing their generation~\cite{autoGenKern}, the vast majority of kernels in well-known libraries such as CUDA\,\footnote{\url{https://docs.nvidia.com/cuda/}} and BLAS~\cite{blas} are still handwritten by experts. 
However, to the best of our knowledge, there are no optimized libraries specifically tailored for transformer execution on edge devices.

On the other hand, given that Attention layers mainly exploit \glspl{GEMM} and Linear layers, \gls{SotA} \gls{DNN} libraries can be reused and extended to execute transformers. Table~\ref{tab:libraryComparison} provides a qualitative description of the existing \gls{DNN} libraries commonly used for TinyML platforms.
XNNPACK is a library commonly used to accelerate high-level machine learning frameworks such as TensorFlow or PyTorch. As indicated in Table \ref{tab:libraryComparison}, it supports a wide range of platforms but does not support multi-core or efficient data reshaping. It is also used in a TinyML context when using the TensorFlow Lite~\cite{tensorflow2015} deployment framework to execute code on Raspberry Pi\,\footnote{\url{https://www.raspberrypi.org/}}. XNNPACK has been expanded to optimize efficiency further by targeting \SI{8}{bit} integer quantized neural networks, which are now the SotA networks in terms of efficiency.

CMSIS-NN~\cite{lai2018cmsis} is the first kernel library for the deployment of \glspl{DNN} that explicitly targets \gls{MCU} class devices, namely platforms relying on ARM Cortex-M cores. CMSIS-NN is tailored for single-core platforms and cannot handle parallelization over any input/output dimensions. It exploits \gls{SIMD} operations on 16 bits when supported, and its main focus is on optimizing the data reuse in the register file to minimize the additional load/store latency, which was observed to be one of the sources of significant overhead.
PULP-NN~\cite{garofalo2020pulp} is another kernel library aiming to perform inference of \glspl{DNN} on RISC-V \glspl{MCU} efficiently. It targets explicitly \gls{PULP} \glspl{SoC} such as GAP9 based on the RV32 XpulpV2 \gls{ISA}. Similarly to CMSIS-NN, it supports \gls{SIMD} and Loop Unrolling, boosting the computational intensity for quantized workloads and minimizing memory access stalls. However, both libraries lack support for fused data marshaling that has been demonstrated to be worth 20\,\% of the latency of transformer networks execution~\cite{ivanov2020data}. Hence, they struggle to efficiently execute operators involving data-layout reconfiguration, such as Transposition. 

The four libraries described above have a few limitations when boosting the performance of transformers' execution. For instance, they do not consider the specific attention layer topology, they do not provide flexible parallelization dimension, and they do not fuse data marshaling and transpose operators, thus missing crucial optimization opportunities. Finally, they do not provide efficient softmax implementation, a key operator in \gls{MHSA}.
To the best of the authors' knowledge, we introduce the first attention-tailored library for \glspl{MCU}. Our library addresses the limitations of the \gls{SotA} \gls{DNN} libraries by providing an optimized parallelization scheme and loop ordering, alleviating the need for expensive data marshaling operations.

\section{Methods}
\label{sec:method}
This section describes our end-to-end deployment toolchain depicted in Figure~\ref{fig:flow}; this framework unlocks the execution of Transformer networks on several \glspl{MCU}. We first give an overview of every tool we use; then, we detail our contributions at each step of the toolchain. Finally, we introduce the three Transformers that we deploy to benchmark our method.

\begin{figure}[t]
    \centering
    \includegraphics[width=\columnwidth]{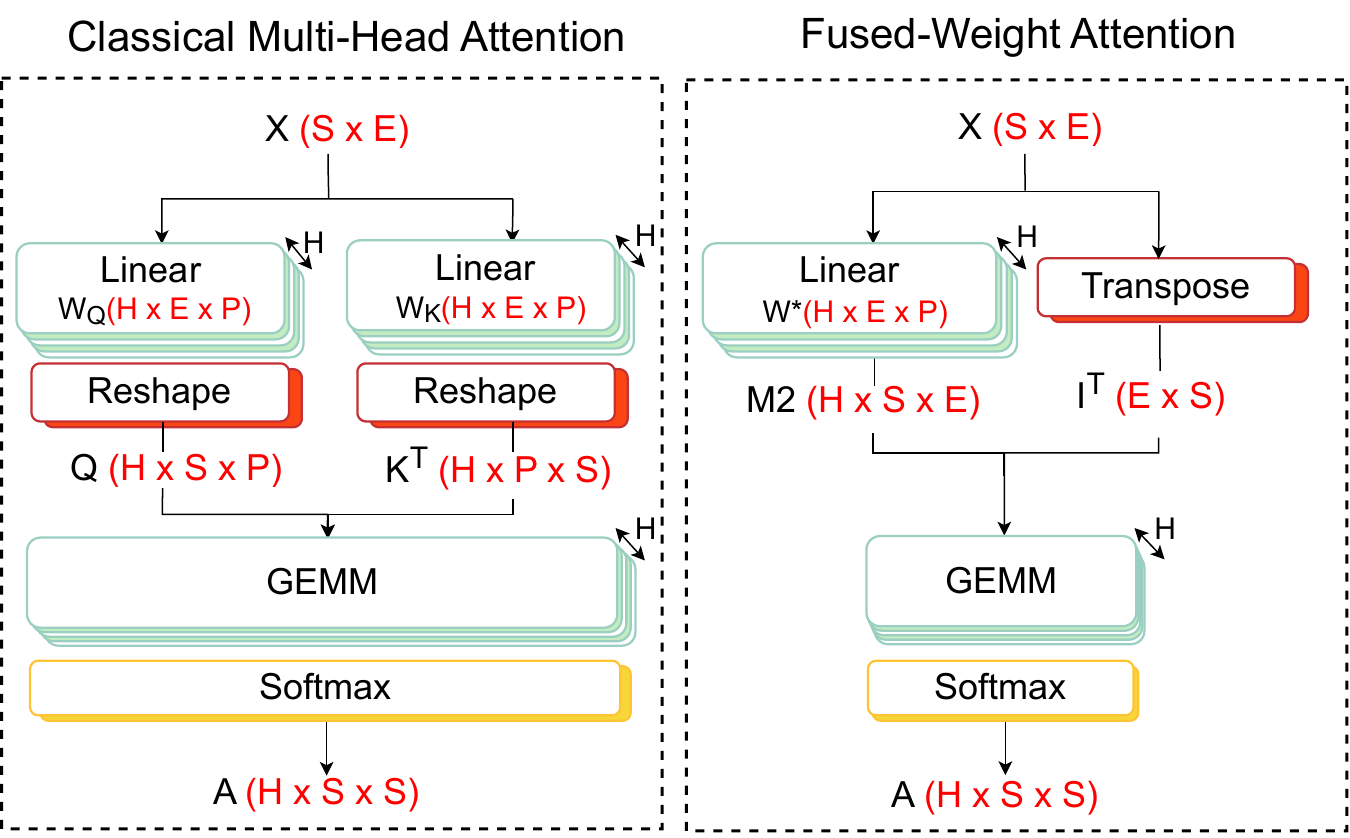}
    \caption{Diagram of the Classical Attention used in the \gls{MHSA} and the proposed Fused-Weight Attention. The dimension of each tensor is specified in red.}
    \label{fig:FW_diagram}
    \vspace{-0.5em}
\end{figure}

\subsection{Deployment Toolchain Overview}
\label{ssec:toolchain_overview}

The input of our toolchain is a PyTorch Transformer floating point model \circledGreen{A} represented as an \gls{ONNX} graph. Then, this graph goes through the optional \gls{FWSA} transformation \circledGreen{B} before being processed by QuantLib\,\footnote{\url{https://github.com/pulp-platform/quantlib}} \circledGreen{C}, our open-source library for model quantization and graph topology transformation. QuantLib's output is a quantized \gls{ONNX} graph compatible with the downstream deployment tool. The deployment frameworks are enhanced by our kernel library \circledGreen{D} and a \gls{DFT} scheme for \gls{MHSA} \circledGreen{E}; their output is code that can be executed on the target platforms \circledGreen{F}. 

We exploit two different deployment tools depending on the target platform. For commercial ARM-based \glspl{MCU} such as STM32, we utilize DumpO, an internally developed automated code generation tool leveraging CMSIS-NN or our optimized transformer library as backend kernels.
GAP9 and other \gls{PULP} platform \glspl{SoC} require additional deployment steps, such as tensor tiling, as they employ an explicitly managed memory hierarchy instead of conventional caches. 
For this reason, in this work, we rely on and enanched DORY~\cite{burrello2021dory}, a \gls{SotA} open-source tool used to deploy \glspl{DNN} on \glspl{MCU} with software-managed caches such as GAP9.

\begin{figure}[t]
    \centering
    \includegraphics[width=\columnwidth]{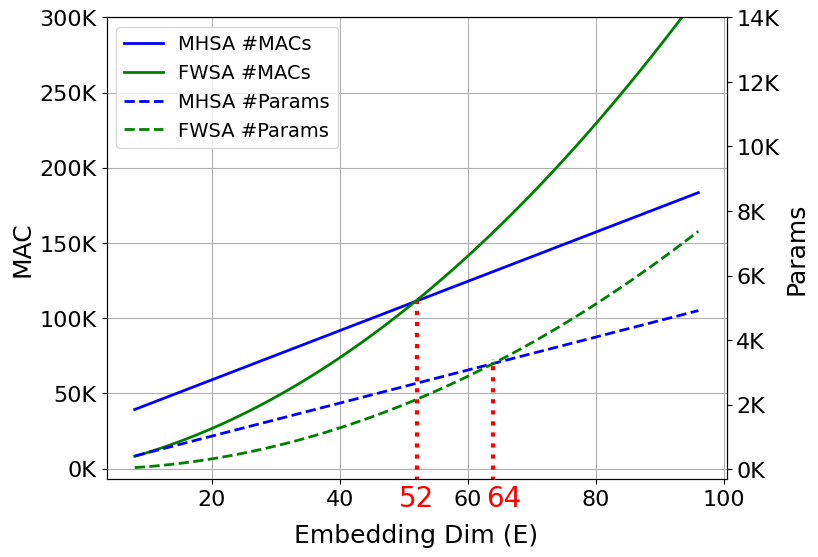}
    \caption{Number of parameters and \gls{MAC} as a function of the embedding dimension $E$ and for $S = 32$, $P = 32$ and $H = 8$. The intersection points happen at $E=52$ and $E=64$ for the number of \gls{MAC} and the number of parameters, respectively.}
    \label{fig:lambda}
    \vspace{-0.5em}
\end{figure}

\subsection[]{Fused-Weight Self-Attention \circledGreen{B}}
\label{ssec:fwsa}

In small-scale Transformers, the embedding size $E$ is often tiny, unlike in larger transformers. We introduce an isomorphic transformation of the \gls{MHSA}, fusing the weights for the $Q$ and $K$ Linear projection offline to exploit this fact. We call this transformation \textit{Fused-Weight Self-Attention}.
This optimization aims to speed up the inference by removing the dependency from $P$ in the number of operations on the \gls{MCU} and the number of parameters to store on-chip. In the equations below, $A$ denotes the attention matrix, $X$ the input, $\phi$ the softmax, and $W_{QK}$ the weights of the $Q$ and $K$ Linear projection. 
We define the fused weight $W^{*}$ such as: $W^{*}=W_Q\cdot W_K^T$
Additionally, $I \in \mathbb{R}^{(S \times E)}$, $W^* \in \mathbb{R}^{(E \times E)}$, $W_{QK} \in \mathbb{R}^{(E \times P)}$, and $A \in \mathbb{R}^{(S \times S)}$.

\begin{equation} \label{eq1}
    A = \phi (X \cdot W_Q \cdot (X \cdot W_K)^T) \\
\end{equation}
\begin{equation} \label{eq2}
    A = \phi (X \cdot W_Q \cdot W_K^T \cdot X^T) \\
\end{equation}
\begin{equation} \label{eq:fwsa}
    A = \phi(X\cdot W^{*}\cdot X^T) \\
\end{equation}

In the equations above, we reorder the operations using the associativity of the matrix product and the transposition's reversal order of the product. The fused weight matrix $W^*$ can be computed offline and considered as fixed parameters during the inference. In detail, we reduce the number of matrix multiplications to perform online from three to two. Figure~\ref{fig:FW_diagram} shows the computational graph of the \gls{FWSA} and the classical Attention.

To evaluate when this transformation is beneficial, we compare the number of operations $O$ and parameters $\theta$ between the \gls{FWSA} and the \gls{MHSA} in Eq.~\ref{eq:numOp}\&\ref{eq:numParam} below:

\begin{equation} \label{eq:numOp}
    \begin{cases}
        O_{MHSA} = 2HSPE + HS^2P \\
        O_{FWSA} = HSE^2 + HS^2E \\
    \end{cases}
\end{equation}

\begin{equation} \label{eq:numParam}
    \begin{cases}
        \theta_{MHSA} = 2HPE \\
        \theta_{FWSA} = HE^2 \\
    \end{cases}
\end{equation}

Looking at Eq.~\ref{eq:numOp}\&\ref{eq:numParam}, we notice that using the \gls{FWSA} is indeed beneficial for some conditions of $S$, $P$, and $E$ values. More specifically, in Eq.~\ref{eq6} and Eq.~\ref{eq7}, we provide the inequalities that need to be satisfied to benefit from the \gls{FWSA} in terms of number of operations and parameters, respectively.

\begin{equation} \label{eq6}
    E < P - \frac{S}{2} + \frac{\sqrt{4P^2 + S^2}}{2} \\
\end{equation}
\begin{equation} \label{eq7}
    E < 2P \\
\end{equation}

In Figure~\ref{fig:lambda}, we show the relationship between $E$ and the number of \gls{MAC}/Parameters for the \gls{MHSA} and \gls{FWSA}.
For $E < 52$, \gls{FWSA} always shows a lower number of \gls{MAC} and parameters.
When considering the intermediate condition of $52 \leq E < 64$, the \gls{FWSA} reduces the number of parameters but increases the number of \glspl{MAC}. Finally, in the third case ($E \geq 64$), the \gls{MHSA} is more advantageous than the \gls{FWSA}.

\subsection[]{Transformers Quantization with QuantLib \circledGreen{C}}

To quantize Transformers, we employ and extend the open-source quantization library QuantLib. QuantLib transforms the \gls{ONNX} graph representing the floating-point model of a \gls{DNN} to an integerized \gls{ONNX} graph. This transformation from a floating-point to an integerized graph is done in two steps.

All operation nodes in the graph, such as matrix multiplication, convolution, or softmax, are replaced by their integer equivalent. As the softmax, \gls{GELU}, and layer normalization present in Transformers do not have an integer equivalent, we extend QuantLib to support the integer version of these operands from I-BERT.

Then, we add a re-quantization step after operations whose output is not represented in \SI{8}{\bit}. For instance, the output type of an integer \gls{GEMM} is \textit{int32}; thus, to convert it to \SI{8}{\bit}, we have to re-quantize it.
We use the uniform symmetric quantizer from the \gls{SotA} \gls{TQT} method~\cite{jain2020tqt}. Below, we describe the functional behavior of the \gls{TQT} quantizer denoted $\psi$: 
\begin{equation}
    \psi(x) = \text{max}(2^{b-1}, \text{min}((x\cdot\epsilon_{mul})>>\epsilon_{div}, 2^{b-1}-1)
\end{equation}
The number of bits of the quantized values is $b$, while $\epsilon_{mul}$ and $\epsilon_{div}$ are the re-quantization parameters.
This quantizer constrains the scale factor to a powers-of-2 (i.e.,  $2^{\epsilon_{div}}$) and uses per-tensor scaling of activations and weights, making it hardware-friendly as we can use the bit-shift operation to apply $\epsilon_{div}$.
Additionally, we use \gls{QAT} to adjust weights to reduce the accuracy loss induced by quantization. 
We note that while \gls{PTQ} is generally used to quantize large-scale Transformers due to their expensive training, in the case of Tiny Transformers, \gls{QAT} is affordable and mitigates the accuracy loss due to quantization.

After applying the necessary transformations and obtaining an integerized \gls{ONNX} graph, we integrate into QuantLib a new export module to format the graph such that it can be ingested by the deployment framework (DORY or DumpO).

\begin{listing}[t]
\centering DumpO-generated pseudocode for ARM \glspl{MCU}
\begin{minted}[mathescape=true,
              linenos,
              escapeinside=||,
              highlightlines={4,5,6,7},
              numbersep=5pt,
              gobble=2,
              frame=lines,
              fontsize=\footnotesize,
              framesep=2mm]{python}
Inputs: X; Output: Y
kernel_params = {param1, param2, etc...};
Kernel(X, Y, kernel_params);
\end{minted}
\centering DORY-generated pseudocode for GAP9
\begin{minted}[mathescape=true,
              linenos,
              escapeinside=||,
              highlightlines={4,5,6,7},
              numbersep=5pt,
              gobble=2,
              frame=lines,
              fontsize=\footnotesize,
              framesep=2mm]{python}
Inputs: X; Output: Y
kernel_params = {param1, param2, etc...};
for (i = 0; i < N|$_{tile}$|; i++)
    DMA_wait(L1|$_{X\_next}$|); swap(L1|$_{X\_next}$|, L1|$_{X\_current}$|); 
    DMA_transfer_async(L1|$_{X\_next}$| <- L2|$_{X}$|[i]);
    Kernel(L1|$_{X\_current}$|, L1|$_{Y\_current}$|, kernel_params);
    DMA_wait(L1|$_{Y\_previous}$|); swap(L1|$_{X\_current}$|, L1|$_{X\_previous}$|); 
    DMA_transfer_async(L2|$_{Y}$|[i] <- L1|$_{Y\_previous}$|);
 \end{minted}
 \caption[]{Example of code generated by DORY and DumpO for executing a \gls{DNN} layer. The input and output tensors are noted as $X$ and $Y$, respectively. The code generated by DORY features tiling with double-buffering, it uses the \gls{DMA} core to fetch input tiles (noted $L1_{X}$) and send back output tiles (noted $L1_{Y}$) asynchronously to the execution of the kernel.} 
 \label{fig:pseudocode_tiling}
 \end{listing}

\vspace{-1em}
\subsection{Deployment Framework Integration}
\label{ssec:depl_framework_integration}

For ARM platforms, we use DumpO to compile the integerized \gls{ONNX} graph into C code. DumpO first transforms the graph so that each node can be executed by one kernel from the available libraries. Then, it parses the graph to match every node with the appropriate kernel and generates the code for the target platform.
We modify the kernels to use the appropriate intrinsics for vector packing and \gls{SIMD} calls to leverage our optimized library in ARM platforms.

This simple code generation strategy does not work for the GAP9 target due to the nature of the hardware architecture.
As mentioned in Sec.~\ref{ssec:toolchain_overview},  unlike ARM processors, the family of GAP processors does not have hardware-managed data caches, meaning that software must handle data movement and tiling. 
While increasing the complexity of the deployment framework, this choice unlocks several optimizations, such as tiling and double-buffering. An example of implementing these optimizations is provided in Listing~\ref{fig:pseudocode_tiling}, comparing the generated code between the ARM/DumpO and GAP/DORY cases.

For this reason, we employ and extend a different tool, DORY, when targeting GAP9. Similarly to DumpO, DORY transforms the graph to fuse operands and features a parser to match layers to kernels. Furthermore, it supports multi-level layer-wise tiling with double-buffered \gls{DMA} transfers to overlap data movement behind computation~\cite{burrello2021dory}. 
The tile dimensions are chosen using constrained optimization~\cite{burrello2021dory} and relies on handcrafted heuristics such as maximizing the L1 memory utilization. DORY also integrates the PULP-NN library~\cite{garofalo2020pulp} in its backend natively.

The baseline version of DORY focuses exclusively on convolutional~\glspl{DNN} and cannot handle Transformers. Furthermore, the baseline DORY targets feed-forward Convolutional~\glspl{DNN}, allowing only one skip connection branch. 
For example, it can support a network such as MobileNet V2~\cite{howard2017mobilenets}, but not SSD~\cite{liu2016ssd}. 
We extend DORY by adding support for Transformer-related operators such as \gls{MHSA} \gls{GELU}, layer normalization, and softmax, adding our specialized library of efficient \gls{MHSA} and \gls{FWSA} kernels \circledGreen{D} as a new backend for DORY.
Additionally, we integrated support for a new tiling scheme \circledGreen{E} and added support for multiple skip connection branches. In the following subsections, we describe in detail this tiling scheme tailored for \gls{MHSA} for MCUs without hardware-managed caches and our optimized kernel library.

\vspace{-1em}
\subsection[]{MHSA Depth-First Tiling \circledGreen{E}}
\label{ssec:dft}

Tiling is usually done layer per layer; the intermediate activation tensors are entirely moved from one level to the other after each \gls{DNN} layer.
This approach is referred to as \emph{\gls{LWT}} and is by far the most common one for deployment at the edge~\cite{burrello2021dory, mei2020zigzag}. 

On the other hand, recently, \emph{\glsreset{DFT}\gls{DFT}} has been gaining traction. The idea behind this is to tile several layers together, effectively using Layer-Wise Tiling but on a group of layers~\cite{mei2023defines, alwani2016fusedCNNAccel}.
The advantage of \gls{DFT} compared to \gls{LWT} is three-fold. First, it reduces the number of memory transfers to the lowest level of the memory hierarchy. Then, it groups compute-bound and memory-bound kernels together, helping to balance the computation vs. communication ratio of the tiles. Finally, it can drastically reduce the peak memory during execution by removing the need to reconstruct a large intermediate feature map.
However, the design space of \gls{DFT} schemes is challenging: the number of combinations of layers potentially tiled together grows exponentially with the number of layers to tile together; furthermore, the more layers one tile together, the more constrained the search space will be for this specific solution. This makes automating the search for good \gls{DFT} scheme hard.

We propose a \gls{DFT} scheme specifically tailored for \gls{MHSA} on \glspl{MCU} and inspired by the flash-attention approach \cite{dao2023flashattention}. This scheme aims to reduce the memory peak usually encountered when reconstructing the attention matrix $A$. More precisely, it tiles the two \glspl{GEMM} of the \gls{MHSA} stage and the softmax operator together. We also force output stationarity in the intermediate tensor to avoid overhead, such as accumulating partial outputs. Hence, the inputs of our \gls{DFT} scheme for a single head are the matrices $K$ and $V$ and $x$ rows of $Q$. Its output is $x$ rows of the $M1$ matrix. The L1 memory size required to apply this tiling scheme, noted $\textit{Mem}_{\text{DFT}}$, can be expressed as follow:

\begin{equation}
    \label{eq:dft_min_size}
    \textit{Mem}_{\text{DFT}}(x) = (2P + S)x + 2PS
\end{equation}
where $Px$ bytes are used to store a tile of $Q$, $Sx$ to store the intermediate tile of $A$, and $Px$ bytes for the output tile of $M$. $2PS$ bytes are used to store a head of $Q$ and $K$.
When we cannot fit L1 memory given the smallest possible $x$, \gls{LWT} is used as a fallback solution.
The result section compares the LWT and DFT approaches in all evaluation scenarios.

\begin{figure}[t]
    \centering
    \includegraphics[width=\columnwidth]{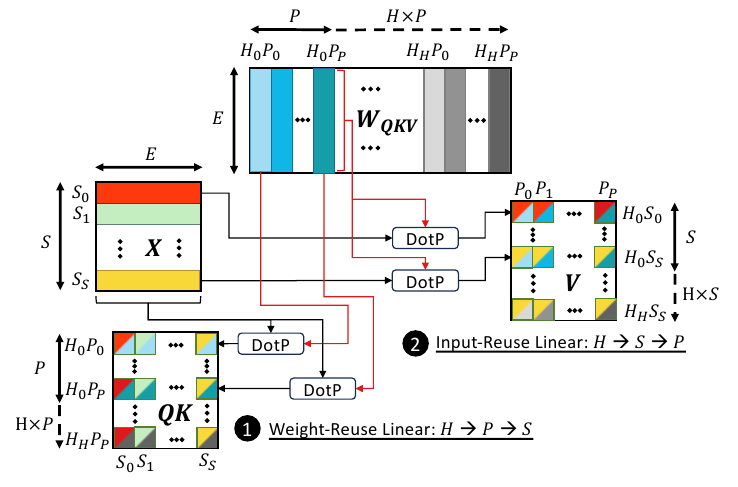}
    \caption{Linear layer dataflow for generating Q, K, and V. The output data layout is $HPS$ and $HSP$. Matrices are filled from top left to bottom right.}
    \label{fig:linear}
    \vspace{-1em}
\end{figure}

\begin{figure}[t]
    \centering
    \includegraphics[width=\columnwidth]{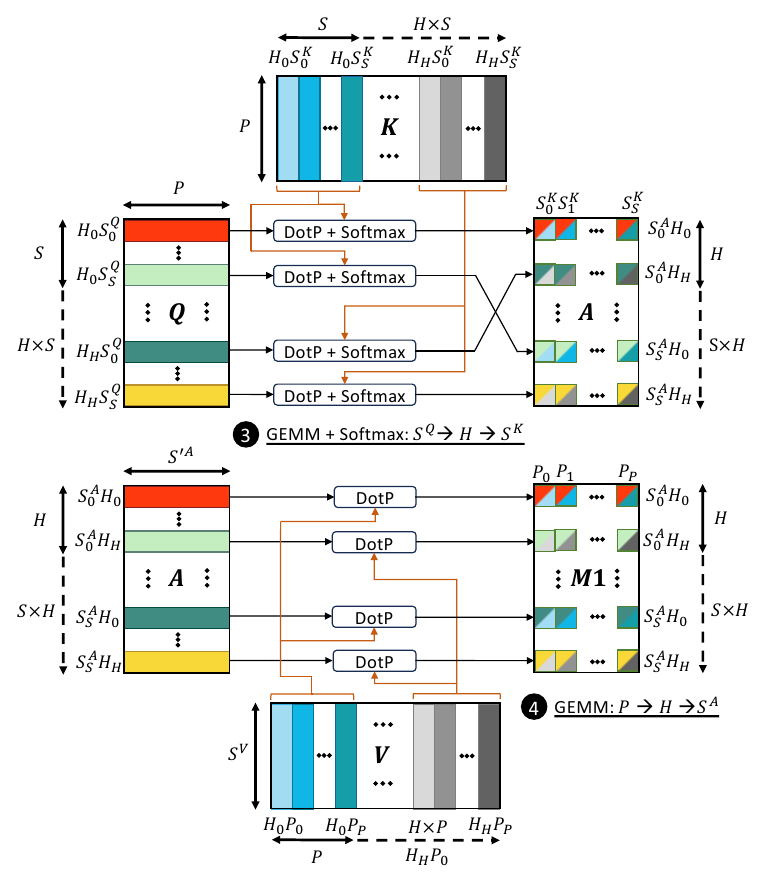}
    \caption{Dataflow of the matrix multiplication kernels designed to work with multiple heads with different data layouts.}
    \label{fig:matmul}
    \vspace{-0.5em}
\end{figure}

\subsection[]{Multi-Head Self-Attention Kernel Library \circledGreen{D}}
\label{ssec:kernel_lib}

Every layer of the \gls{MHSA}, described in Figure~\ref{fig:attention}, is followed by transpositions, reshape, or concatenation operations. These memory marshaling operations are hard to execute efficiently on multi-core processors as they only comprise data movement. Hence, hiding the data transfer latency behind computation is impossible. Therefore, their impact on performances is non-negligible, causing overheads up to 20\,\% in some network execution \cite{ivanov2020data}.
Furthermore, \gls{SotA} \glspl{DNN} libraries~\cite{garofalo2020pulp, lai2018cmsis} do not tune data layout to avoid data marshaling operations and do not fine-tune their parallelization scheme for \gls{MHSA}.
Hence, they perform poorly for executing \gls{MHSA}, exhibiting low data reuse and sub-linear scaling on multi-core processors.
We propose our optimized library of tailored kernels for each layer in the \gls{MHSA} to address these issues.
Each kernel of the library embeds the re-quantization step and takes \SI{8}{\bit} input and output tensors. The scaling factors and the bias of the linear projection are stored on \SI{16}{bit}.

\subsubsection{Kernel Execution Loop}
\label{subsec:kernelExecLoop}

We follow three primary guidelines for kernel optimizations: i) keep the parallelization (when available on the target platform) as much as possible on the $H$ dimension; ii) exploit output stationarity; and iii) produce the output tensors sequentially (i.e., element $i + 1$ in the innermost dimension is always generated immediately after the $i$-th element).

When the memory constraints of the platform allow it, we parallelize our kernels on the $H$ dimension. Spatially unrolling the heads has several advantages. First, this dimension is present in every kernel of the \gls{MHSA}. Second, it allows for minimizing synchronization among cores since the heads perform independent computation, avoiding race conditions. Note that this parallelization scheme is not chosen on high-performance devices such as \glspl{GPU}, given the high number of cores compared to the number of heads. 

We also support the typical scenario, where the kernels are tiled over $H$, and then we parallelize over the outermost dimension other than $H$. Both kernel variants are available in the library, allowing the developer to choose the appropriate parallelization dimension depending on the constraints of the platform and the hyper-parameters of the \gls{MHSA}.

The second guideline saves memory by avoiding the storage of many intermediate \texttt{int32} accumulators for the partial outputs, as described in~\cite{burrello2021dory}.
Besides the memory saving, output stationarity maximizes the exploitation of the dot-product \gls{SIMD} instructions, as demonstrated in~\cite{garofalo2020pulp}.

Finally, our motivation to enforce the last guideline is to prevent potential computational overhead induced by the additional operations in the innermost loop to compute storage locations.

\subsubsection{Linear Layers}
\label{sssec:linear}

\begin{listing}[tb]
\begin{minted}[mathescape=true,
              linenos,
              escapeinside=||,
              highlightlines={4,5,6,7},
              numbersep=5pt,
              gobble=2,
              frame=lines,
              fontsize=\footnotesize,
              framesep=2mm]{python}
Inputs: I, W; Outputs: Q, K, V
C = H / CORES
|H$_{start}$| = |\textcolor{darkspringgreen}{\textbf{min}}|(C * |CORE$_{ID}$|, H); |H$_{end}$| = |\textcolor{darkspringgreen}{\textbf{min}}|(H|$_{start}$| + C, H)
  |\textcolor{blue}{\textbf{LH}}|: for (h = |H$_{start}$|; h < |H$_{end}$|; h++)  
    |\textcolor{blue}{\textbf{LP}}|: for (p = 0; p < P/2; p++) 
      |\textcolor{blue}{\textbf{LS}}|: for (s  = 0; s < S/4; s++) 
        S0...7 = {0};
        |\textcolor{blue}{\textbf{LE}}|: for (e = 0; e < E/4; e++) 	
          A1 = I(4s); A2 = I(4s + E); 
          A3 = I(4s + 2E); A4 = I(4s + 3E);	
          B1 = W(hpE + 2pE); B2 = I(hpE + (2p+1)E);
	  S0 += |\textcolor{darkspringgreen}{\textbf{sdot4}}|(A1,B1); S1 += |\textcolor{darkspringgreen}{\textbf{sdot4}}|(A1,B2); 
	  S2 += |\textcolor{darkspringgreen}{\textbf{sdot4}}|(A2,B1); S3 += |\textcolor{darkspringgreen}{\textbf{sdot4}}|(A2,B2); 
	  S4 += |\textcolor{darkspringgreen}{\textbf{sdot4}}|(A3,B1); S5 += |\textcolor{darkspringgreen}{\textbf{sdot4}}|(A3,B2); 
	  S6 += |\textcolor{darkspringgreen}{\textbf{sdot4}}|(A4,B1); S7 += |\textcolor{darkspringgreen}{\textbf{sdot4}}|(A4,B2); 
	O(h,2p,4s) = |\textcolor{darkspringgreen}{\textbf{quant}}|(S0);
	O(h,2p,4s+1) = |\textcolor{darkspringgreen}{\textbf{quant}}|(S1);  
	O(h,2p,4s+2) = |\textcolor{darkspringgreen}{\textbf{quant}}|(S2);
	O(h,2p,4s+3) = |\textcolor{darkspringgreen}{\textbf{quant}}|(S3);  
	O(h,2p+1,4s) = |\textcolor{darkspringgreen}{\textbf{quant}}|(S4);
	O(h,2p+1,4s+1) = |\textcolor{darkspringgreen}{\textbf{quant}}|(S5);  
	O(h,2p+1,4s+2) = |\textcolor{darkspringgreen}{\textbf{quant}}|(S6);
	O(h,2p+1,4s+3) = |\textcolor{darkspringgreen}{\textbf{quant}}|(S7); 
  
 \end{minted}
 \caption[]{Example of kernel pseudocode of sub-layer \circled{1}.} 
 \label{fig:pseudocode_layer}
 \end{listing}

Figure~\ref{fig:linear} depicts two distinct implementations for the Linear layers, used to project the input to the $Q$, $K$, and $V$ matrixes and compute the output of the \gls{MHSA}.
Implementation \circled{1}, is called \gls{WRL} and is used to project the $\mathbf{V}$ tensor from $\mathbf{X}$, while we use the implementation \circled{2}, named \gls{IRL}, for $\mathbf{Q}$ and $\mathbf{K}$. These two implementations have a different loop ordering and data layout.

From a functional point of view, for a single head, one can write the functions of the \gls{WRL} and \gls{IRL} kernels as follows:
\begin{equation} \label{eq:linear} 
    \begin{cases}
        \varphi_{IRL}(X, W^T) = O \\ 
        \varphi_{WRL}(X, W^T) = O^T \\
    \end{cases}
\end{equation}
Where $\varphi$ is the function of the kernel, $X$, $W$, and $O$ are the Input, Weight, and Output tensors, respectively.

WRL kernel produces output data in the $HPS$ order, allowing the subsequential matrix multiplication to ingest data sequentially without stridden access. In this way, we remove the data-reshaping operator, reducing the overall number of memory accesses. At every iteration of the $P$ loop, the whole input ($S\times E$) matrix is multiplied by a single weight sample ($1\times E$).
From outermost to innermost, we order the loop as $H\rightarrow P\rightarrow S\rightarrow E$.
On the other hand, the \gls{IRL} requires the output layout to be $HSP$ to allow the following matrix multiplications to read data sequentially. Therefore, the loop order in this case is $H \rightarrow S \rightarrow P \rightarrow E$, from outermost to innermost.
Contrary to WRL, at every iteration of \gls{IRL}, the $S$ loop, a single input sample ($1\times E$), is multiplied by a weight-head ($E\times P$).

The last Linear layer, which projects the output tensor of the matrix multiplication, noted $M$, to the final output, uses the $S \rightarrow E \rightarrow H\times P$ loop order. Since the $H$ dimension is a part of the reduction dimension, we parallelize this kernel over $S$, the outermost loop.

Listing~\ref{fig:pseudocode_layer} reports an example of the pseudocode of layer~\circled{1} with the \texttt{RV32IMCXpulpV2} \gls{ISA} and GAP9 target.
In the innermost loop, we exploit the \texttt{sdot4} operator to perform 4 \gls{MAC} operations in a single instruction. 
Additionally, we perform loop unrolling in the same fashion as~\cite{garofalo2020pulp}. We select the window size of the loop unrolling to be $4 \times 2$, meaning that we perform 8 dot-products on 8 register-allocated accumulators at each loop iteration. This optimization allows us to avoid \gls{RAW} hazards and increase the register files data reuse.
Similarly to PULP-NN~\cite{garofalo2020pulp}, incrementing the number of produced output values in a single iteration, e.g., to 16, would cause an increase in the number of required registers to 24 (16 for outputs, 4 for inputs, 4 for weights); we observe that the compiler\footnote{A precompiled RISC-V toolchain for GAP IoT Application Processor compiled with gcc-9.4.0 available at https://github.com/GreenWaves-Technologies/gap-riscv-gnu-toolchain} can not generate valid code, in this case, keeping all accumulators allocated to registers: some of them are spilled to the stack to make room for operands, causing extra load/store operations and reducing the overall performance.
Conversely, reducing the number of registers employed causes a reduction in the \gls{MAC}/load ratio and impairs the performance.

\subsubsection{Matrix Multiplications}

To optimize matrix multiplications, we optimally order the loop executions and parallelize over the outermost dimensions to improve performance.
Figure~\ref{fig:matmul} reports two different implementations for the two \glspl{GEMM} in the \gls{MHSA} kernel. 

The Matmul-Softmax \circled{3} fuses the matrix multiplication with the integer softmax; it uses the $S \rightarrow H \rightarrow S$ loop order; $P$ is the dimension over which the reduction is performed.
Each iteration of the $H$ loop produces a new row of $A$. We apply the softmax to the produced row (e.g., the first one, $S_0 H_0$).
The final Linear layer of the \gls{MHSA} (see Figure~\ref{fig:attention}) performs the reduction over the concatenated dimension of $H$ and $P$. Hence, to avoid this concatenation operator, we choose the output data layout to be $SHS$, interleaving the heads with the sequence length.
For instance, the first row of $Q$ ($S \times 1$) is multiplied with the first head of $K$ ($P \times S$), then the softmax is applied, resulting in the first row of $A$ ($1 \times S$). Next, the process is repeated with the second head of $K$, resulting in the second row of $A$. Once the first row of $Q$ has been matched with the whole $K$ matrix, we reiterate using the next row of $Q$.

Matmul \circled{4} produces the tensor $M$, which is then fed to the output projection Linear layer.
Its implementation is straightforward given the design of the previous layers \circled{1} and \circled{3}. The loop execution order is $S \rightarrow H \rightarrow P$, with $P$ as the innermost dimension. The reduction dimension is the innermost $S$ of the $\mathbf{M1}$ matrix. 

\subsubsection{Fused-Weight Self-Attention Kernel}

Figure~\ref{fig:FW_diagram} shows the computational graph of the \gls{FWSA}, and Sec.~\ref{ssec:fwsa} describes its equation. 
Like the other kernels of our library, the \gls{FWSA} kernel uses \gls{SIMD} and loop unrolling to maximize hardware utilization.
The \gls{FWSA} kernel is broken down into two parts, one responsible for the computation of $M2$ and the other for $A$.

The kernel to compute $M2$ is derived from the Linear \gls{IRL} kernel \circled{1} used for the Linear projection of $Q$ and $K$ and described in detail in Sec.~\ref{sssec:linear}. Its loop ordering is $H \rightarrow S \rightarrow E$ to create output with the $HSE$ layout. The second part of the kernel is inspired by the Softmax-Matmul \circled{3}. It reuses the same loop ordering except that its reduction dimension is $E$ instead of $P$. The output layout of the \gls{FWSA} kernel is the same $SHS$ data layout as \circled{3}.

\subsection{End-to-end Tiny Transformers Applications}
\label{ssec:applications}

    To demonstrate the performances of our deployment toolchain, we decided to use three real-world Tiny Transformer networks targeting very different applications, from biosensing wearables use cases like arrhythmia classification or seizure detection to more traditional hand gesture recognition.

    Two of the three transformers we are benchmarking in this paper have been introduced in~\cite{busia2022eegformer}, demonstrating better performance than the \gls{CNN} counterpart. The other one, TR-Former, has been first introduced in~\cite{burrello2021mcuIsAllYouNeed}. In the following subsections, we summarize the rationale and describe the architecture of TR-Former, EEGFormer, and ECGFormer.
    
    Recent research has empirically shown that Transformers can tolerate a quantization down to \SI{4}{\bit}~\cite{frantar2023gptq}. However, quantizing to \SI{4}{bit} would negatively impact execution efficiency for all the targeted \glspl{MCU} due to the lack of hardware support for \SI{4}{bit} \gls{SIMD}. The only benefit would be to divide by two the memory footprint compared to \SI{8}{bit} quantization. Because of the performance impact and potential accuracy degradation, we use a \SI{8}{\bit} quantization scheme.

    \begin{figure}[t]
      \centering
        \includegraphics[width=\columnwidth]{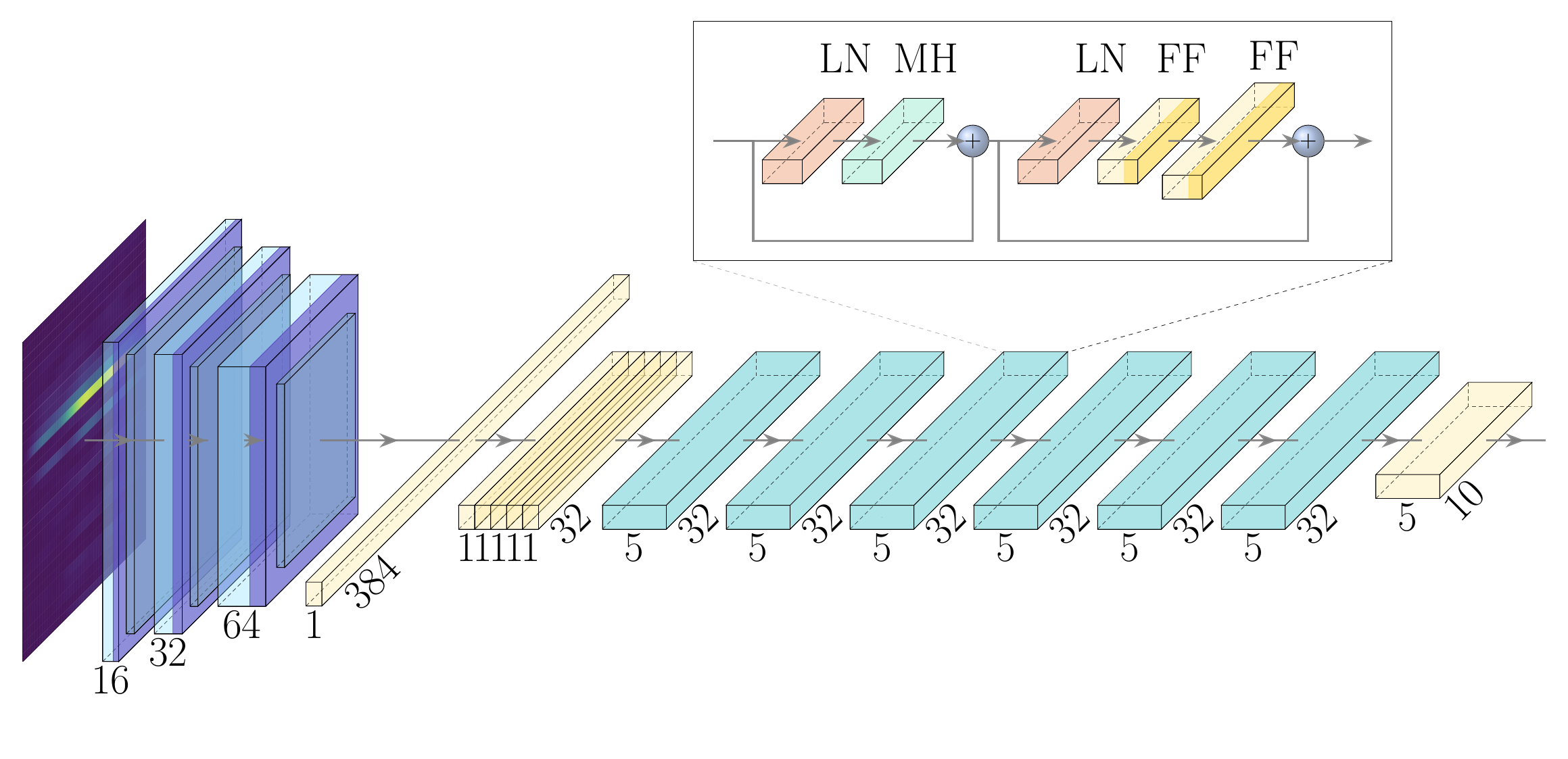}
      \caption{Overall architecture of TR-Former. The front end comprises 3 Tokenizer blocks, each containing Pointwise Convolution, Depthwise Convolution, and Pooling, followed by a Linear layer. Each of the six encoder blocks consists of layer normalization layers (LN), a \gls{MHSA} layer (MH), and Linear layers (FF).}
      \label{fig:TRFormerArchitecture}
    \end{figure}
    
    \subsubsection{Hand Gestures Recognition: TR-Former}
    
    The first application is hand-gesture recognition based on short-range radar. We use the Tiny Transformer proposed in~\cite{burrello2021mcuIsAllYouNeed} and trained on the TinyRadar dataset. It demonstrates the feasibility and the advantages of Transformers on TinyML platforms.
    The dataset consists of over 10K recordings of 11 hand gestures by 26 people made with a short-range radar.
    The architecture is shown in Figure~\ref{fig:TRFormerArchitecture}.
    First, they use a \gls{CNN} stage with 3 blocks of convolutions, each followed by a batch norm, ReLU6 activation layer, and pooling. The input resolution is reduced at each block while the number of output channels scales from 16 to 64. A final Linear layer projects the input sample to the embedding dimension $E = 32$. 
    Using $S=5$ processed input time samples as a sequence, a $5\times32$ input is fed to the Transformer backend.
    The 6 layers constituting the backend are identical to the ones of~\cite{dosovitskiy2020image}, with $S = 5$, $E = 32$, $P = 32$, and $H = 8$.
    Finally, the output of the encoder is fed to a dense layer, which is used as a classifier, returning a prediction for each time step.

    \subsubsection{Seizure Detection: EEGFormer}

    The second Tiny Transformer targets a non-obtrusive and non-stigmatizing \gls{EEG} acquisition setup to detect seizures.
    EEGFormer is a network derived from BioFormers and is trained on the CHB-MIT dataset. Its first stage features two layers of 1D convolutions, each followed by batch normalization. This convolutional stage transforms a \gls{EEG} signal of 2048 samples to a $81 \times 32$ matrix used as an input for the second stage of the network.
    This second stage is a traditional transformer encoder containing a \gls{MHSA} layer of the following dimensions: $S = 81$, $E = 32$, $P = 32$, and $H = 8$.
    A detailed description of the architecture can be found in~\cite{busia2022eegformer}.
    
    \subsubsection{Arrhythmia Classification: ECGFormer}

    The last task we target is arrhythmia classification; the dataset used is MIT-BIH. ECGFormer's~\cite{busia2024ECGFormer} architecture is also based on BioFormer and targets ultra-low power applications. Compared to EEGFormer, the main differences are the use of the RR interval in the final classification stage, the substantially smaller number of samples (198) in the input signal, and the number of convolutions in the embedding stage (from two for EEGFormer and one for ECGFormer. Additionally, ECGFormer features a single transformer encoder block with \gls{MHSA} dimensions of $S = 66$, $E = 16$, $P = 2$, and $H = 8$. More details on the architecture can be found in~\cite{busia2024ECGFormer}.

    \begin{figure}[t]
    \vspace{-1em}
    \centering
    \includegraphics[trim={0.8cm 0cm 0cm 0cm}, width=\columnwidth]{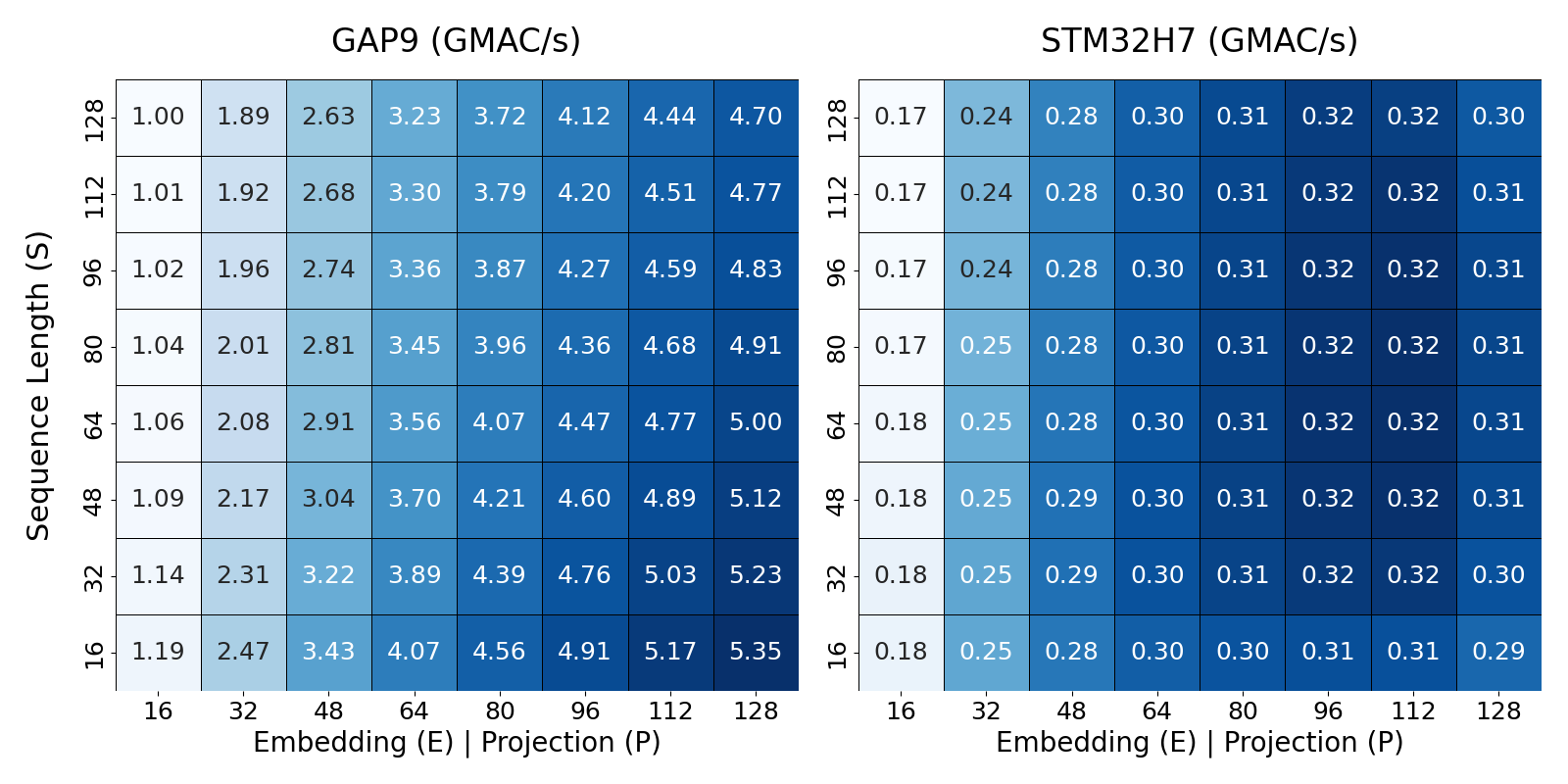}
    \caption{Heatmap representing the throughput of our \gls{MHSA} kernel on GAP9 and STM32H7 for various input dimensions of $(S \times E)$. The projection dimension $P$ is equal to $E$ and the number of heads ($H$) is set to 8.}
    \label{fig:kernelHeatmap}
    \vspace{-0.5em}
\end{figure}

\vspace{-1em}
\section{Results And Discussion}
\label{sec:results}
In this section, we detail our evaluation setup, and we benchmark our library against the \gls{SotA} libraries, CMSIS-NN for ARM platforms and PULP-NN~\cite{lai2018cmsis, garofalo2020pulp} for RISC-V ones, to showcase the advantage of our method for various input dimensions, number of cores, and hardware platforms. Afterward, we demonstrate the usage of the library with the deployment tools to deploy three end-to-end \gls{SotA} transformers for different edge applications. Different input dimensions and architectural hyperparameters characterize each application.
To conclude, in Sec.~\ref{ssec:ablation}, we provide an ablation study of the impact of the \gls{FWSA} and the \gls{DFT} on latency and memory footprint on these applications. 

\vspace{-0.5em}
\subsection{Evaluation Setup}

We benchmarked and compared our library with \gls{SotA} kernels on the three platforms introduced in Sec.~\ref{subsec:hardware}.
On GAP9, we use internal performance counters to measure the cycles of both single attention blocks and the entire Transformers execution.
Furthermore, to simulate larger dimensions of the GAP9 internal memories, we rely on the event-driven simulator GVSoC from GreenWaves Technologies, which enables cycle-accurate simulations.
To measure the power consumption, we use Nordic Semiconductor's Power Profiler Kit 2 (PPK2)\,\footnote{\url{https://www.nordicsemi.com/Products/Development-hardware/Power-Profiler-Kit-2}} with a sampling frequency of \SI{100}{\kilo\hertz}. We measure the power of the cluster and fabric controller and the off-chip \gls{RAM} and Flash. We consider a hot start, meaning that we neglect the movements of weights from L3 to L2, executed a single time for consecutive inferences.

For STM32L4 and STM32H7, we use internal hardware counters to measure the number of cycles, and we consider constant power consumption measured under an identical workload to estimate the energy.
For STM32H7, we activate the data and instruction caches and store the weights and activation tensors in the \gls{SRAM} to maximize the performance. For the STM32L4 platform, as we do not have caches in this processor, we store every constant tensor (input and weights) in the read-only data section of the Flash.

\begin{figure}[t]
\vspace{-1em}
  \centering
    \includegraphics[width=\columnwidth]{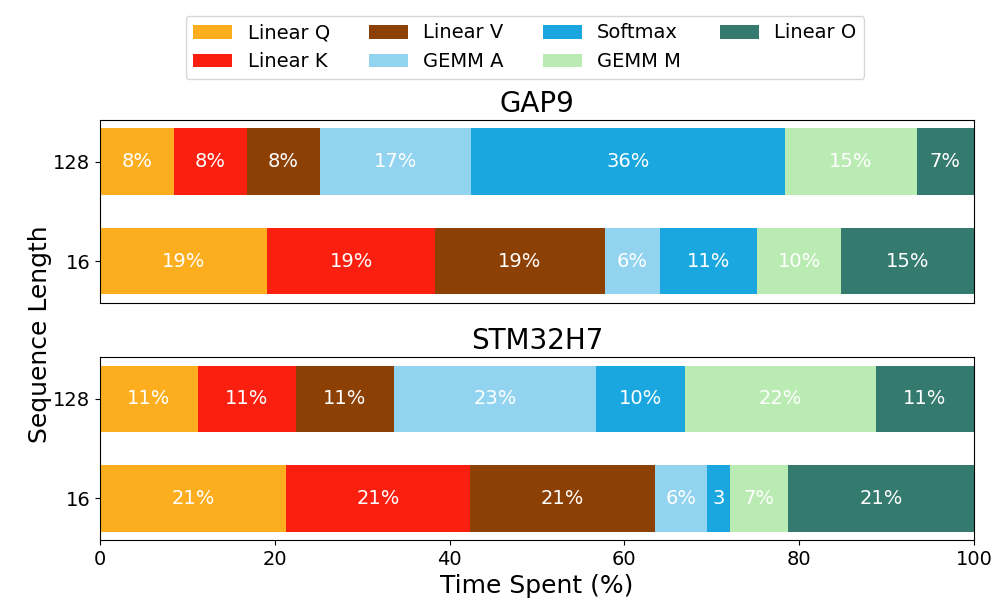}
  \caption{Breakdown of the execution time of each MHSA layer on GAP9 and STM32H7 for sequence lengths of 16 and 128. The other dimensions are fixed to $E=64$, $P=64$, and $H=8$.}
  \label{fig:layer_breakdown}
  \vspace{-1.5em}
\end{figure}
\begin{figure}[t]
\noindent
\begin{minipage}{\columnwidth}
    \includegraphics[width=\columnwidth]{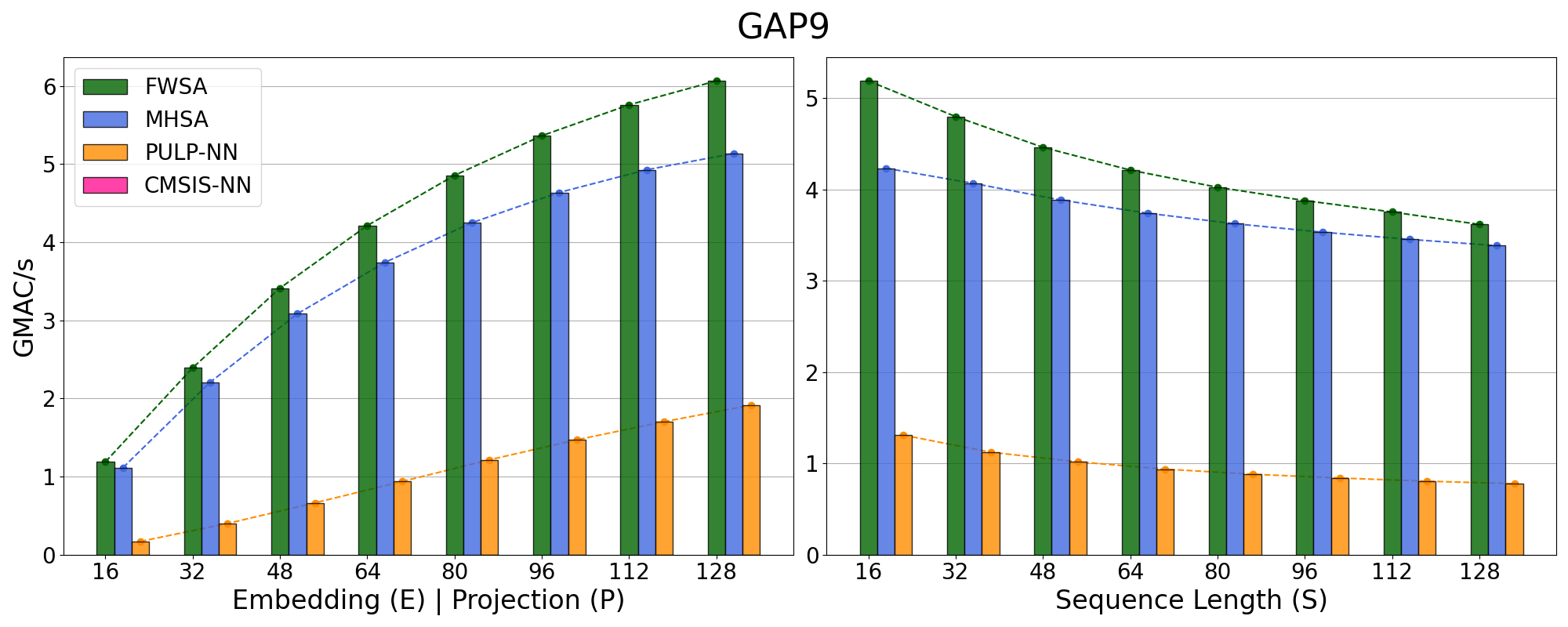}
    \includegraphics[width=\columnwidth]{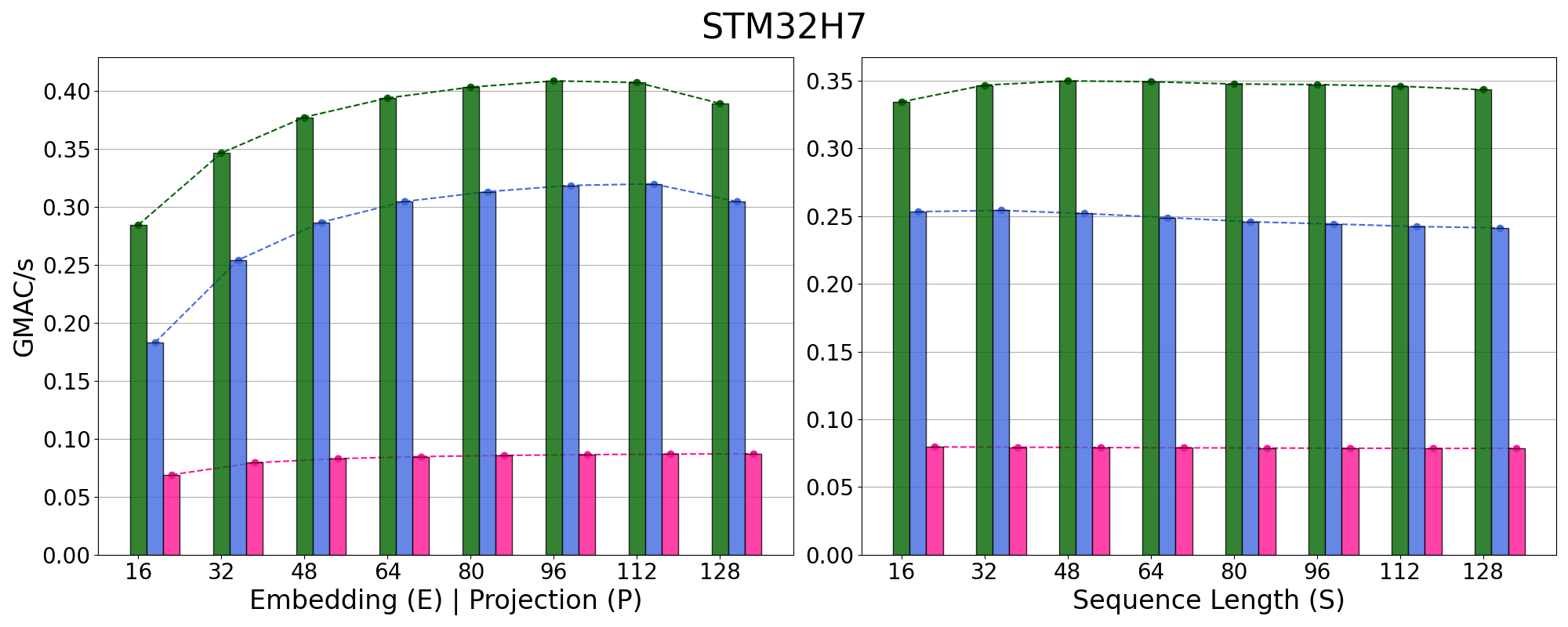}
    \captionof{figure}{Thoughtput comparison of attention block on GAP9 and STM32H7 for various embedding sizes and sequence lengths. We use our \gls{MHSA} and \gls{FWSA} kernels and \gls{SotA} kernel libraries PULP-NN or CMSIS, depending on the platform.}
    \label{fig:macDimSE}
    \vspace{-1em}
\end{minipage}
\end{figure}

\begin{figure}[t]
    \centering
    \includegraphics[trim={0cm 0cm 0cm 2cm}, width=\columnwidth]{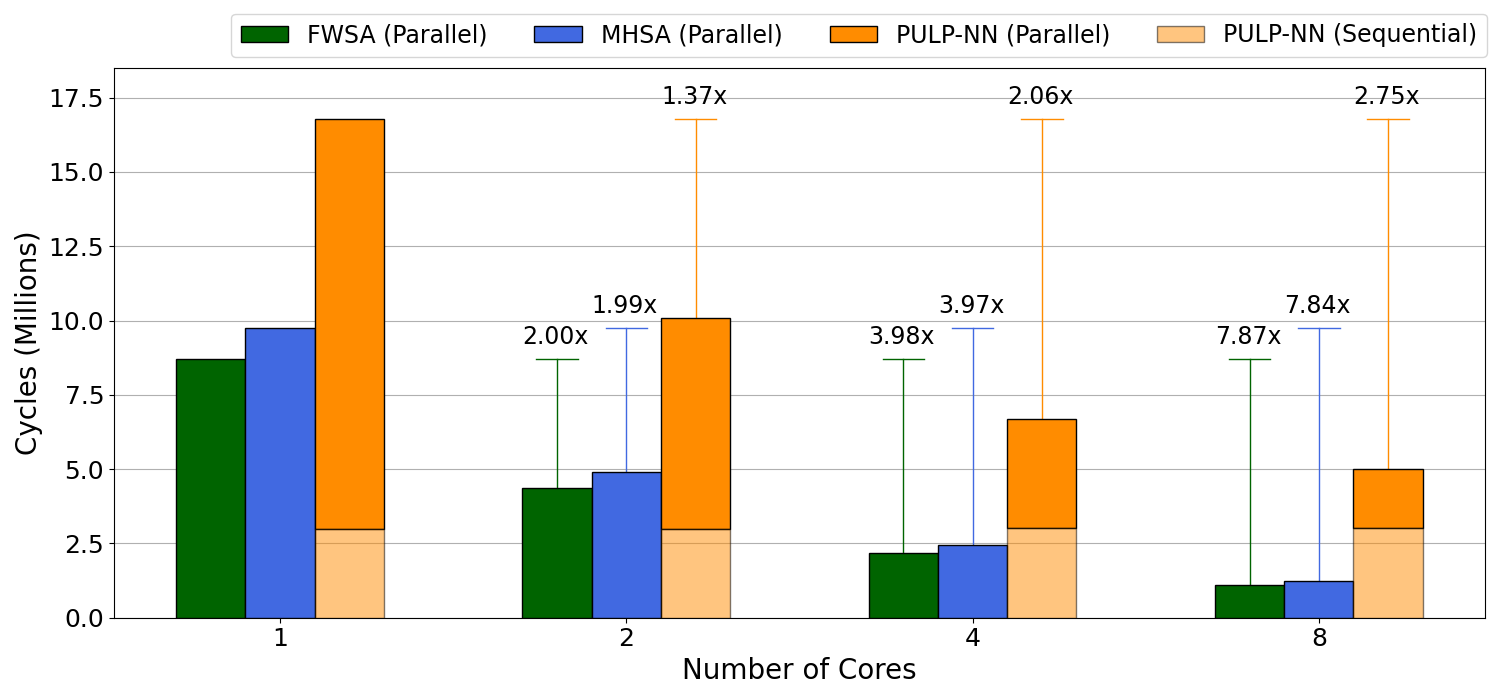}
    \caption{Parallelization of the \gls{MHSA} on GAP9 for the three \gls{SotA} libraries. For PULP-NN, we indicate the breakdown of sequential and parallel execution time. Fused-Weight and Vanilla are completely parallelized. The dimensions of the \gls{MHSA} are ($S=64$, $E=64$, $P=64$, and $H=8$).}
    \label{fig:kernelParallelization}
    \vspace{-0em}
\end{figure}
\begin{table}[t]
\centering
\renewcommand{\arraystretch}{1.35}
\caption{Performances of end-to-end applications on GAP9 at maximum frequency and at the most energy-efficient configuration.}
\begin{tabular}{|c|p{0.7cm}p{0.8cm}|p{1.1cm}p{0.8cm}|p{1.1cm}p{0.8cm}|} \cline{4-7}
\multicolumn{1}{c}{} & \multicolumn{2}{p{1.5cm}|}{} & \multicolumn{2}{p{1.9cm}|}{370MHz/50mW} & \multicolumn{2}{p{1.9cm}|}{230MHz/20mW} \\ \hline
Task                 & MACs     & Cycles     & Lat. (\si{\ms}) & E. (\textmu \si{\joule)} & Lat. (\si{\ms}) & E. (\textmu \si{\joule)} \\ \hline
\multicolumn{1}{|l|}{EEG} &  7.35M   & 3.33M      &     9.42     &  460        & 13.76        &   310                \\ \hline
\multicolumn{1}{|l|}{ECG} &  0.97M   & 1.15M      &     2.85     &  120        & 4.28         &    90                \\ \hline
\multicolumn{1}{|l|}{TR}&   6.00M    & 1.92M      &     5.49     &  315        &     8.52      &           207         \\ \hline
\end{tabular}

\label{tab:endToEnd}
\end{table}

\vspace{-0.5em}
\subsection{Micro-benchmarking: MHSA and FWSA}
\subsubsection{Input Size Scaling}
\label{ssec:input_scaling}
First, we show the performance of our library for various dimensions of the input tensor $X$. To benchmark kernel optimizations on GAP9 without considering data movements, we increase the L1 memory size using GVSoC and directly store both weights and activations at this level.
We report only STM32H7 performance for the ARM-based platforms, given that the same conclusions can be drawn for the STM32L4, which shares the same \gls{ISA} and architecture.

Figure~\ref{fig:kernelHeatmap} shows the performance in terms of GMAC/s when executing a \gls{MHSA} with different input sizes. On the $x$-axis, we increase $E|P$ dimensions, while on the $y$-axis, we increase the $S$ dimension. The number of heads has been kept constant at 8 to maximize parallelization on GAP9 and at 1 for the STM32H7.
In the figure, we observe two significant trends for GAP9: firstly, a notable decrease in efficiency occurs with an increase in sequence length; secondly, an improvement in efficiency when $E$ and $P$ increase.

The first effect is due to the latency of the softmax nonlinearity. We measure the complexity of the \gls{MHSA} with the number of \glspl{MAC} of the Linear and \gls{GEMM} layers. Thus, the softmax counts as zero \glspl{MAC} but strongly impacts the overall latency.
The latency of softmax rises proportionally with the dimension of the vector on which it is computed, $S$.
The effect of the softmax on performance for large sequences is quantified in Figure~\ref{fig:layer_breakdown}. For both platforms, the proportion of time spent on the softmax drastically increases by a factor of 3.3\,$\times$ when going from a sequence length of 16 to 128.

The efficiency growth with $E$ and $P$ is well visible in Figure~\ref{fig:kernelHeatmap}: up to 4.7\,$\times$ better between $E|P=16$ and $E|P=128$ with $S=128$. This is because $P$ and $E$ are the dimensions over which the reduction is made in five out of the six matrix multiplications of the \gls{MHSA}. Since we produce the output tensors sequentially, as explained in subsection \ref{subsec:kernelExecLoop}, increasing the reduction dimension leads to better utilization of the \gls{SIMD} units and less overhead due to loop indexes and pointer computations.
For the STM32H7 platform, when we increase $E|P$, we observe the same behavior as for GAP9: the throughput is 1.72\,$\times$ higher for $E|P=128$ than for $E|P=16$ with $S=64$.
We notice that an increase of $S$ leads to a reduction in performance only for low $E|P$ values. When $E$ is higher, the softmax operation's latency on STM32H7 is negligible compared to the other blocks, and, therefore, it does not negatively impact the throughput.

Figure~\ref{fig:macDimSE} reports $E|P$ scaling with constant $S$, and $S$ scaling with constant $E|P$ on GAP9 and STM32H7 for our \gls{MHSA} and \gls{FWSA} kernels, comparing them with the two \gls{SotA} kernel baselines, i.e., PULP-NN and CMSIS-NN.
It can be noticed that our kernels (\gls{MHSA} and \gls{FWSA}) consistently show higher throughput than PULP-NN and CMSIS-NN on both platforms. 
On GAP9, for the various values of $S$ and $E|P$, the average throughput of the \gls{MHSA} and \gls{FWSA} kernels is 4.04\,$\times$ and 4.53\,$\times$ higher than the PULP-NN implementation. In STM32H7, compared to CMSIS-NN, the throughput is 3.28\,$\times$ and 4.45$\,\times$ higher on average for the \gls{MHSA} and \gls{FWSA}, respectively.

\subsubsection{Parallelization Scaling on GAP9}
\label{ssec:parallelization}

\begin{figure}[t]
  \vspace{-1em}
  \centering
    \includegraphics[width=\columnwidth]{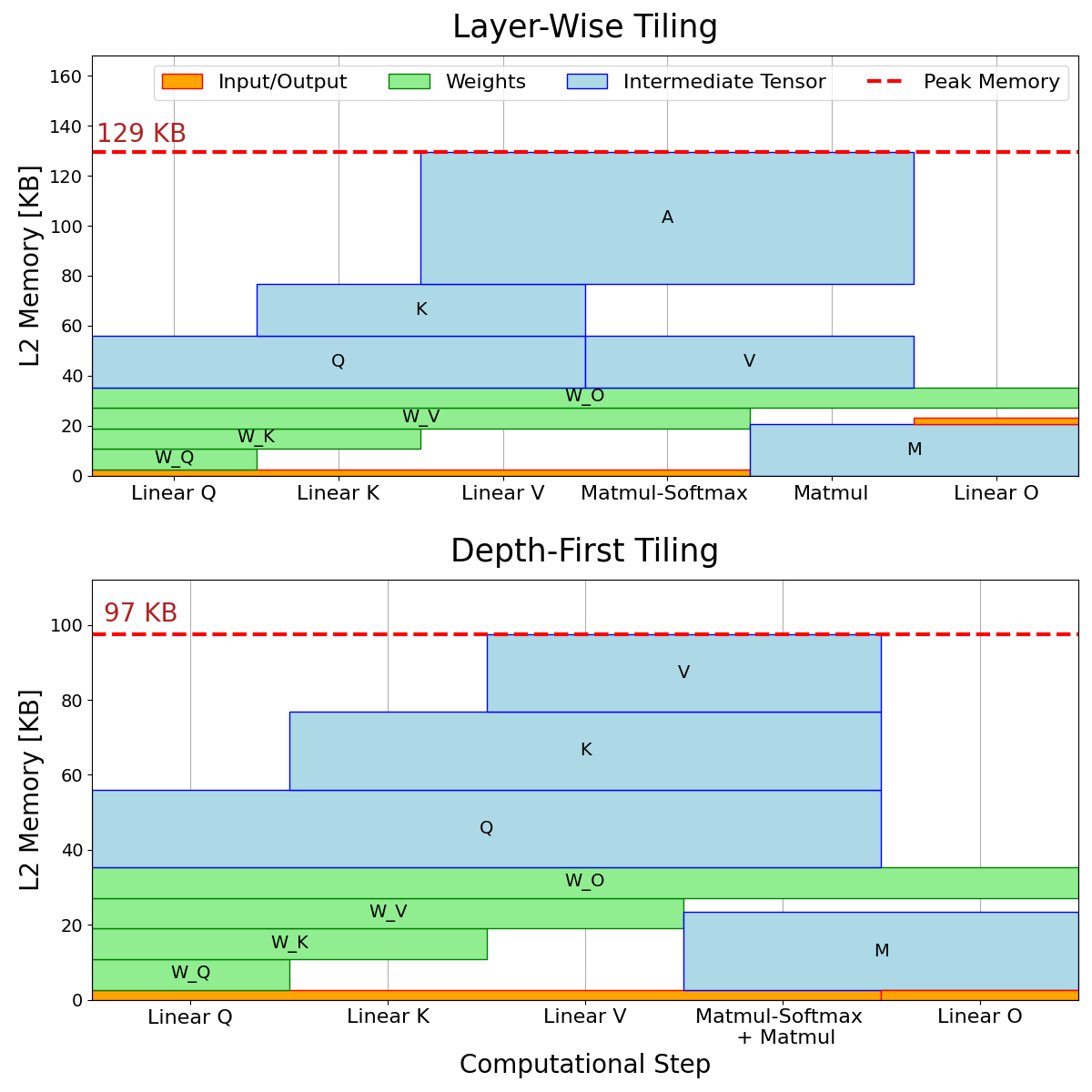}
  \caption{L2 memory allocation breakdown for each layer of the \gls{MHSA} for Classical and Depth-First tiling schemes. The dimensions of the \gls{MHSA} are from EEGFormer \cite{busia2022eegformer} ($S=81$, $E=32$, $P=32$, and $H=8$). Peak L2 memory utilization is indicated in red.}
  \label{fig:tiling}
  \vspace{-0em}
\end{figure}

\begin{table}[t]
\centering
\renewcommand{\arraystretch}{1.35}
\caption{Ablation study of the effect of Fused-Weight attention and Depth-First Tiling on the runtime and L2 memory peak}
\begin{tabular}{|c|c|c|c|c|c|} \cline{2-6}
  \multicolumn{1}{c|}{} &                      \multicolumn{3}{c|}{Memory Peak}    & \multicolumn{2}{c|}{Cycles} \\ \cline{2-6} 
  \multicolumn{1}{c|}{} &  \multicolumn{2}{c|}{MHSA}         &  FWSA    & \multirow{2}{*}{MHSA} &  \multirow{2}{*}{FWSA} \\ \cline{1-4} 
  Tasks & LWT & DFT & LWT &             &               \\ \hline 
  EEG  &       129.3 KB  &  \textbf{97.1} KB &        121.2 KB &    1.07M    &\textbf{1.01M} \\ \hline 
  ECG   &        39.0 KB  &   \textbf{6.3} KB &         38.5 KB &\textbf{553K}&      569K     \\ \hline 
  TR   &        34.2 KB  &          34.2 KB  &\protect\textbf{24.9} KB &   52K       &\textbf{34K}   \\ \hline 
\end{tabular}
\label{tab:optComparison}
\end{table}

\begin{table*}[t]
\centering
\caption{Comparison of our kernel library with PULP-NN and CMSIS-NN onto three commercial MCUs. For each application, we only report our fastest kernel.}
\renewcommand{\arraystretch}{1.35}
\begin{tabular}{@{}ccccccc@{}}    
\hline
\multicolumn{1}{@{}l|}{Platform}                     & \multicolumn{2}{l|}{GAP9}           & \multicolumn{2}{l|}{STM32L4}    & \multicolumn{2}{l@{}}{STM32H7}      \\
\multicolumn{1}{@{}l|}{Core(s)}                      & \multicolumn{2}{l|}{9 RISC-V}       & \multicolumn{2}{l|}{1 Cortex-M4}    & \multicolumn{2}{l@{}}{1 Cortex-M7}      \\
\multicolumn{1}{@{}l|}{Power (mW) / Frequency (MHz)} & \multicolumn{2}{l|}{50mW / 370MHz}   & \multicolumn{2}{l|}{10mW / 80MHz} & \multicolumn{2}{l@{}}{234mW / 480MHz} \\ \hline
\multicolumn{1}{@{}l|}{Kernels}                      & \multicolumn{1}{c}{Ours} & \multicolumn{1}{l|}{PULP-NN~\cite{lai2018cmsis}} & Ours & \multicolumn{1}{l|}{CMSIS-NN~\cite{lai2018cmsis}} & \multicolumn{1}{c}{Ours} & \multicolumn{1}{c}{CMSIS-NN~\cite{lai2018cmsis}}  \\ \hline
\multicolumn{7}{@{}l}{EEGFormer MHSA (S=81 , E=32 , P=32, H=8)}                                                                                             \\ \hline
\multicolumn{1}{@{}l|}{Cycles}                       & 1.01M& \multicolumn{1}{c|}{5.85M }&  39.71M  & \multicolumn{1}{c|}{60.17M}&  13.6M     &    36.47M   \\
\multicolumn{1}{@{}l|}{Time/Inference (ms)}          &  2.84& \multicolumn{1}{c|}{15.96 }&  496.37  & \multicolumn{1}{c|}{752.12}&  28.33     &    75.98    \\
\multicolumn{1}{@{}l|}{Energy (\textmu \si{\joule)}} &  136 & \multicolumn{1}{c|}{ 683  }&  4963.8  & \multicolumn{1}{c|}{7521.3}&  662.98    &    1777.87  \\
\multicolumn{1}{@{}l|}{MACs/cycle}                   &  5.94& \multicolumn{1}{c|}{ 1.03 }&  0.15    & \multicolumn{1}{c|}{0.10}  &  0.44      &    0.16     \\
\multicolumn{1}{@{}l|}{Throughput (GMAC/s)}          &  2.20& \multicolumn{1}{c|}{ 0.38 }&  0.012   & \multicolumn{1}{c|}{0.0079}&  0.21      &    0.079    \\
\multicolumn{1}{@{}l|}{Energy Efficiency (GMAC/s/W)} & 46.09& \multicolumn{1}{c|}{ 8.88 }&  1.21    & \multicolumn{1}{c|}{0.79}  &  9.07      &    3.38     \\ \hline
\multicolumn{7}{@{}l}{ECGFormer MHSA (S=66 , E=16 , P=2, H=8)}                                                                                              \\ \hline
\multicolumn{1}{@{}l|}{Cycles}                       &553K  & \multicolumn{1}{c|}{3.21M }&  3.28M   & \multicolumn{1}{c|}{5.75M} &  1.77M     &    2.98M    \\
\multicolumn{1}{@{}l|}{Time/Inference (ms)}          & 1.63 & \multicolumn{1}{c|}{8.70  }&  41      & \multicolumn{1}{c|}{71.87} &  3.69      &    6.08     \\
\multicolumn{1}{@{}l|}{Energy (\textmu \si{\joule)}} & 62.1 & \multicolumn{1}{c|}{316.0 }&  410     & \multicolumn{1}{c|}{718.75}&  86.28     &    142.34   \\
\multicolumn{1}{@{}l|}{MACs/cycle}                   & 0.37 & \multicolumn{1}{c|}{ 0.06 }&  0.05    & \multicolumn{1}{c|}{0.03}  &  0.10      &    0.06     \\
\multicolumn{1}{@{}l|}{Throughput (GMAC/s)}          & 0.14 & \multicolumn{1}{c|}{ 0.02 }&  0.0043  & \multicolumn{1}{c|}{0.0024}&  0.048     &    0.029    \\
\multicolumn{1}{@{}l|}{Energy Efficiency (GMAC/s/W)} & 3.63 & \multicolumn{1}{c|}{ 0.66 }&  0.43    & \multicolumn{1}{c|}{0.24}  &  2.04      &    1.24     \\ \hline
\multicolumn{7}{@{}l}{TR-Former MHSA (S=5 , E=32 , P=32, H=8)}                                                                                              \\ \hline
\multicolumn{1}{@{}l|}{Cycles}                       & 34K  & \multicolumn{1}{c|}{95K   }&  1.18M   & \multicolumn{1}{c|}{1.71M} &  357K      &    1.04M    \\
\multicolumn{1}{@{}l|}{Time/Inference (ms)}          & 0.14 & \multicolumn{1}{c|}{0.26  }&  14.75   & \multicolumn{1}{c|}{21.37} &  0.74      &    2.16     \\
\multicolumn{1}{@{}l|}{Energy (\textmu \si{\joule)}} &  4.92& \multicolumn{1}{c|}{11.40 }&  147.5   & \multicolumn{1}{c|}{213.73}&  17.4      &    50.70    \\
\multicolumn{1}{@{}l|}{MACs/cycle}                   &  5.19& \multicolumn{1}{c|}{1.85  }&  0.17    & \multicolumn{1}{c|}{0.12}  &  0.58      &    0.20     \\
\multicolumn{1}{@{}l|}{Throughput (GMAC/s)}          &  1.92& \multicolumn{1}{c|}{0.68  }&  0.014   & \multicolumn{1}{c|}{0.010} &  0.28      &    0.095    \\
\multicolumn{1}{@{}l|}{Energy Efficiency (GMAC/s/W)} & 58.51& \multicolumn{1}{c|}{15.60 }&  1.4     & \multicolumn{1}{c|}{0.97}  &  11.89     &    4.08     \\ \hline
\end{tabular}
\label{tab:mcuComparison}
\end{table*}
Figure~\ref{fig:kernelParallelization} details the performance of an attention block using the \gls{MHSA}, the \gls{FWSA}, and the baseline PULP-NN kernels on GAP9 with 1, 2, 4, and 8 cores. As can be noticed, the speed-up of the PULP-NN baseline from 1 to 8 cores is only 2.75\,$\times$ while our kernels reach 7.87\,$\times$ and 7.84\,$\times$ speed-up for \gls{FWSA} and \gls{MHSA}, respectively.  
We identify three main reasons for the better parallelization scaling: (i) unlike PULP-NN, in both our kernels, we parallelize over the outermost loop, requiring fewer synchronization steps. (ii) Moreover, since the \gls{GEMM} in the PULP-NN implementation exploits individual Linear kernels, the cores split the computation on the same dimension on which the softmax has to be executed. In this way, all the cores must be synchronized before the softmax, executed from CORE 0, strongly impacting the parallelization. (iii) Similarly, our approach eliminates the data marshaling operations executed sequentially from CORE 0 inside PULP-NN. 
These last two problems can be observed in Figure~\ref{fig:kernelParallelization}, which shows the sequential part of the kernel in light colors: while it is constant for the PULP-NN baseline, we eliminate it in our kernels.

\vspace{-1em}
\subsection{End-to-end Transformers performance}

Table~\ref{tab:endToEnd} describes the latency and energy of the end-to-end execution of the three Tiny Transformers introduced in Sec. \ref{ssec:applications}. 
We here report the end-to-end performance on GAP9 only, given that we can exploit all the optimizations and features of our kernels, i.e., the influence of parallelization, \gls{FWSA}, and \gls{DFT}. A more detailed analysis of the performance of the attention blocks of these three networks for all hardware platforms is reported in Sec. \ref{sec:soa}.
Here, we report only the results obtained with the best combination of optimizations to minimize latency. In detail, we use the \gls{FWSA} on EEGFormer and TR-Former to reduce the latency, while we use MHSA for  ECGFormer. Additionally, we do not use the \gls{DFT} as it only reduces the L2 memory consumption, and GAP9 features a large enough L2 memory when executing a single transformer. Note that this is given by the specific shapes of the networks of our use cases, where the memory transfers are entirely hidden by computation with double buffering. Therefore, reducing the memory-transfer time does not improve the overall latency of the network.
Further, we show two different hardware configurations: the first one minimizes the latency and runs at \SI{370}{\mega\hertz} with a power consumption of \SI{50}{\mW}. The second configuration runs at \SI{230}{\mega\hertz}, consuming \SI{20}{\mW} and is the most energy-efficient point. 
For the three networks (EEGFormer, ECGFormer, and TR-Former), we obtain a best latency of \SI{9.42}{\ms}, \SI{2.85}{\ms}, and \SI{5.49}{\ms}, respectively, always respecting the real-time constraint imposed by the data acquisition.
The most efficient configuration leads to an energy consumption of \SI{310}{\micro\joule}, \SI{90}{\micro\joule}, and \SI{207}{\micro\joule} for the three networks.

\vspace{-1em}
\subsection{Ablation Study: Optimizations Impact}
\label{ssec:ablation}

In this subsection, we analyze the individual impact of the \gls{FWSA} and the \gls{DFT} for the GAP9 platform.
Table \ref{tab:optComparison} provides an ablation study of the effect of these optimizations on the three attention blocks of the networks detailed in the previous section in terms of memory and latency saved. As said above, given that the \gls{DFT} does not impact the latency, we only report memory saving for this optimization. However, note that in actual application scenarios, such as object detection, multiple networks often run on the same platform. Thus, reducing the memory footprint of each single network is crucial. For the \gls{FWSA}, we measure both the impact on the memory peak and the number of cycles.

Concerning the \gls{FWSA}, its memory peak and the number of operations can be computed offline with Eq.~\ref{eq:numOp}. Using this equation, we find that the number of operations of \gls{FWSA} compared to the \gls{MHSA} is reduced by 11\,\% and 23\,\% for EEGFormer and TR-Former, respectively, while it increases by 30\,\% for ECGFormer. We measure a reduction of the number of cycles of 6\,\% for EEGFormer and 35\,\% for TR-Former. The difference is explained by the hyperparameters of the transformers that influence the efficiency of the single layers and by the softmax, whose operations are not included in the number of operations of Eq.~\ref{eq:numOp}. Reciprocally, for ECGFormer, the number of cycles increases less than expected, by only 3\,\%. This effect is caused by the modification of the reduction dimension of the \gls{GEMM} inside the ECGFormer attention block, from $P=2$ for the \gls{MHSA} to $E=32$ for the \gls{FWSA}, which strongly improves the usage of the \gls{SIMD} and loop unrolling.

Concerning the L2 memory peak, Table~\ref{tab:optComparison} shows that the \gls{FWSA} reduces the memory peak only for TR-Former. Compared to \gls{MHSA}, the \gls{FWSA} does not store the $Q$ and $K$ tensors to generate $A$, effectively skipping the \textit{Linear V} computation step. Therefore, the \gls{FWSA} reduces the overall memory peak of the network only when this step is the most memory-demanding.
Additionally, the table also shows that using \gls{DFT} allows a substantial reduction of the memory peak for the other two networks, 24\,\% for the EEGFormer attention block and 84\,\% for the ECGFormer one. Indeed, our \gls{DFT} scheme aims at avoiding storing the attention matrix $A$ of dimension $(H\times S \times S)$. Hence, \gls{MHSA} with a large ratio between sequence length and projection dimension will benefit more from it. Concerning TR-Former, the memory peak happens at the \textit{Linear V} computation step; therefore, the \gls{DFT} does not reduce it.

Figure~\ref{fig:tiling} details the L2 memory allocation at each step of the sequential and depth-first tiling for the \gls{MHSA} of EEGFormer. The x-axis shows the individual computational steps of the \gls{MHSA} while the y-axis represents the L2 memory allocation. A computational step ends when the tiled output reconstructs the complete output tensor. As can be noticed, \gls{DFT} reduces the number of computational steps by one. As explained in Sec.~\ref{ssec:dft}, this is because we tile the two \glspl{GEMM} of the \gls{MHSA} together and reconstruct only the tensor $M1$.
Consequently, the memory peak moves \SI{129}{\kilo\byte}, to \SI{97}{\kilo\byte} mainly due to skipping the storage of the matrix $A$. 

\vspace{-1em}
\subsection{Comparison with State-of-the-art}
\label{sec:soa}

We demonstrated in Sec.~\ref{ssec:input_scaling} and Sec.~\ref{ssec:parallelization} the higher efficiency and scaling capabilities of our kernels compared to the \gls{SotA}. In this section, we compare them on the three real transformer networks, characterized by layers dimensions often unsuited to exploit kernel efficiency.
For instance, the value of the projection dimension $P$ of ECGFormer is 2. Hence, we cannot use \gls{SIMD} for kernels where $P$ is the most internal loop, such as the first \gls{GEMM} operation (see Fig.~\ref{fig:attention}). 

Table~\ref{tab:mcuComparison} showcases latency, energy, and different efficiency metrics for the attention blocks of our three use cases onto three hardware targets.
We reported the reference \gls{SotA} kernel results for each target compared to our best kernel alternative, i.e., FWSA for EEGFormer and TR-Former and MHSA for ECGFormer. 
Specifically, to compare with \gls{SotA} PULP-NN and CMSIS-NN kernels libraries, we leverage their optimized linear layers kernels, add additional loops, data marshaling operations, and I-BERT's integer softmax~\cite{kim2021bert} to implement the attention layer. 
The average improvement in terms of latency of our best kernels on the three attention blocks is 4.80\,$\times$, 1.57\,$\times$, and 2.43\,$\times$ for GAP9, STM32L4, and STM32H7, respectively. 
Interestingly, the significant improvement compared to the \gls{SotA} for the three different hardware platforms is always associated with a different attention block.

For GAP9, our kernels reach the top latency improvement of 5.80\,$\times$ on EEGFormer where our kernels can maximally exploit the \gls{SIMD} usage on the $P$ dimension and exploit the parallelization also on the $S$ dimension. As discussed in Sec.~\ref{ssec:parallelization}, the speed-up of our library over PULP-NN on GAP9 is primarily due to parallelizing the execution of the softmax and getting rid of sequential data marshaling operations such as transpositions.
For the STM32 platforms, we obtain significant latency gains of 1.73\,$\times$ and 2.91\,$\times$ for STM32L4 and STM32H7 on the ECGFormer and TR-Former attention block, respectively.
Noteworthy, these two networks further highlight the ability of our kernels also to manage \textit{non-ideal} \gls{MHSA} parameters, e.g., P=2 for ECGFormer or S=5 for TR-Former, which, on the other hand, strongly impair the performance of \gls{SotA} kernels. 
Compared to CMSIS-NN, our kernels feature higher data reuse for these single-core platforms thanks to loop reordering and data marshaling operations fusion.

\vspace{-0.3cm}
\section{Conclusion}
\label{sec:conclusion}
In this work, we proposed an end-to-end flow to enable efficient deployment of small Transformer models onto commercial \glspl{MCU}. Our kernel library, tailored for \gls{MHSA}, together with our optimized schedule and tiling strategy, allows us to speed up the execution of the attention block by a factor of 2.94$\times$ on average on RISC-V and ARM platforms. Furthermore, we demonstrate the efficiency of our flow by deploying three Tiny Transformers onto the GAP9 \gls{MCU}, reaching an average energy consumption and latency of \SI{202}{\micro\joule} and \SI{5.92}{\ms}, respectively. Our work is open-source at \url{https://github.com/pulp-platform/pulp-transformer}.

\tiny
\bibliographystyle{IEEEtran}
\bibliography{main}

\vspace{-6em}
\begin{IEEEbiography}[{\includegraphics[width=0.82in,height=0.82in,clip,keepaspectratio]{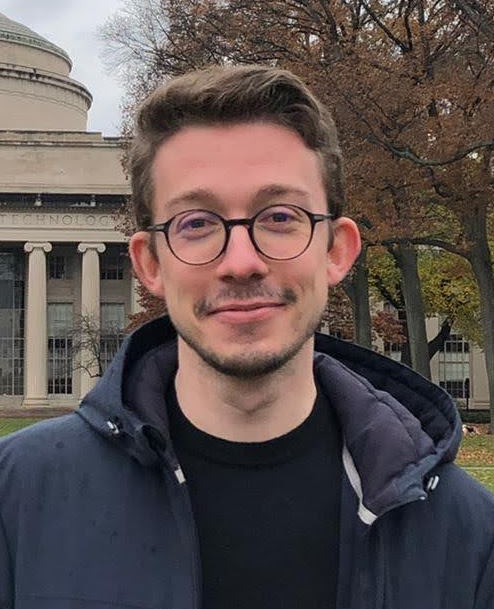}}]{Victor Jean-Baptiste Jung}
 received his Bachelor's Degree in Computer Science and Engineering Physics from Juniata College, and the Master's Degree in Computer Science from the Institut Supérieur de l’Electronique et du Numérique of Lille (ISEN Lille) in 2022. After 3 months as a research intern with KU Leuven’s MICAS Research group supervised by Prof. Marian Verhelst, he started his Ph.D. at the Integrated Systems Laboratory with Prof. Dr. Luca Benini. His current research interests include Efficient deployment of ML models on Microcontrollers, Tiny Transformers, Scheduling and Quantization.
\end{IEEEbiography}
\vspace{-6em}
\begin{IEEEbiography}[{\includegraphics[width=0.82in,height=0.82in,clip,keepaspectratio]{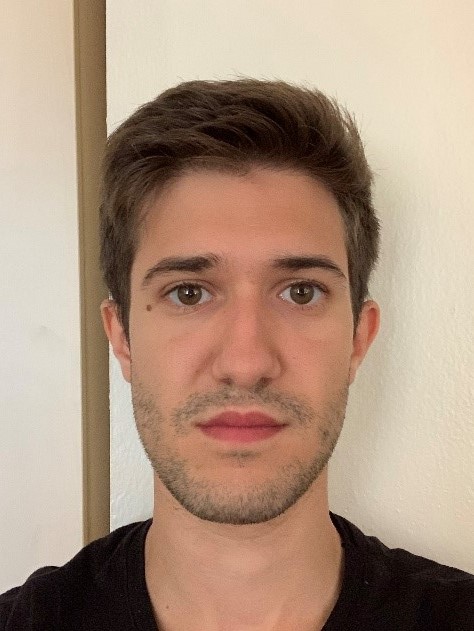}}]{Alessio Burrello}
is currently a research assistant at Politecnico di Torino. He received his M.Sc. and Ph.D. degrees in Electronic Engineering at the Politecnico of Turin, Italy, and the University of Bologna, respectively, in 2018 and 2023. His research interests include parallel programming models for embedded systems, machine and deep learning, hardware-oriented deep learning, and code optimization for multi-core systems. He has published over 70 papers in peer-reviewed international journals and conferences. His work has been awarded different times, including best paper awards at IEEE ISVLSI 2023 and IEEE BioCAS 2018.
\end{IEEEbiography}
\vspace{-6em}
\begin{IEEEbiography}[{\includegraphics[width=0.82in,height=0.82in,clip,keepaspectratio]{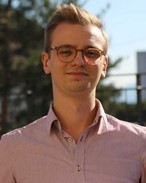}}]{Moritz Scherer} received the B.Sc. and M.Sc. degree in electrical engineering and information technology from ETH Zürich in 2018 and 2020, respectively, where he is currently pursuing a Ph.D. degree at the Integrated Systems Laboratory. His current research interests include the design of ultra-low power and energy-efficient circuits and accelerators as well as system-level and embedded design for machine learning and edge computing applications.
Moritz Scherer received the ETH Medal for his Master’s thesis in 2020.
\end{IEEEbiography}
\vspace{-6em}
\begin{IEEEbiography}[{\includegraphics[width=0.82in,height=0.82in,clip,keepaspectratio]{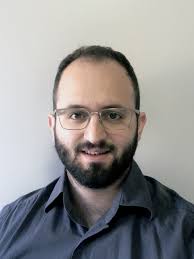}}]{Francesco Conti} (Member, IEEE) received the Ph.D. degree in electronic engineering from the University of Bologna, Italy, in 2016.
He is currently a Tenure-Track Assistant Professor with the DEI Department, University of Bologna. From 2016 to 2020, he held a research grant with the University of Bologna and a Post-Doctoral Researcher with ETH Zürich.
His research is centered on hardware acceleration in ultra-low power and highly energy-efficient platforms, with a particular focus on System-on-Chips for Artificial Intelligence applications.
His research work has resulted in more than 90 publications in international conferences and journals and was awarded several times, including the 2020 IEEE \textsc{Transactions on Circuits and Systems I: Regular Papers} Darlington Best Paper Award.
\end{IEEEbiography}
\vspace{-6em}
\begin{IEEEbiography}[{\includegraphics[width=0.82in,height=0.82in,clip,keepaspectratio]{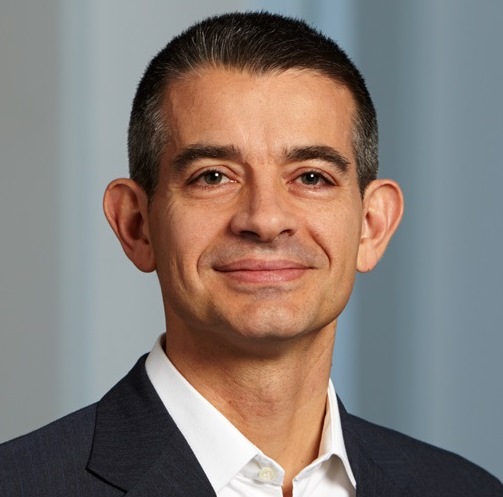}}]{Luca Benini}
is the Chair of Digital Circuits and Systems at ETH Z\"urich and a Full Professor at the University of Bologna.
He has served as Chief Architect for the Platform2012 in STMicroelectronics, Grenoble.
Dr. Benini’s research interests are in energy-efficient systems and multi-core SoC design. 
He is also active in the area of energy-efficient smart sensors and sensor networks. 
He has published more than 1’000 papers in peer-reviewed international journals and conferences, five books and several book chapters. 
He is a Fellow of the ACM and a member of the Academia Europaea.
\end{IEEEbiography}

\end{document}